\documentclass[sigconf,authorversion,nonacm]{acmart}
\makeatletter
\def\@ACM@checkaffil{% Only warnings
    \if@ACM@instpresent\else
    \ClassWarningNoLine{\@classname}{No institution present for an affiliation}%
    \fi
    \if@ACM@citypresent\else
    \ClassWarningNoLine{\@classname}{No city present for an affiliation}%
    \fi
    \if@ACM@countrypresent\else
        \ClassWarningNoLine{\@classname}{No country present for an affiliation}%
    \fi
}
\makeatother

\usepackage{amsfonts}       % blackboard math symbols
\usepackage[utf8]{inputenc} % allow utf-8 input
\usepackage[T1]{fontenc}    % use 8-bit T1 fonts
\usepackage{hyperref}
\usepackage{nicefrac}       % compact symbols for 1/2, etc.
\usepackage{microtype}      % microtypography
\usepackage{xcolor}         % colors
\usepackage{mathtools}
\usepackage{amsthm}
\usepackage{multirow}
\usepackage{microtype}
\usepackage{subfigure}

\theoremstyle{plain}

\theoremstyle{definition}

\theoremstyle{remark}

\def\r{\mathbf{r}}

\def\h{\mathbf{h}}
\def\m{\mathbf{m}}

\def\g{\mathbf{g}}
\def\x{\mathbf{x}}

\def\z{\mathbf{z}}

\def\R{\mathbb{R}}

\begin{document}

\title{Deep Integrated Explanations}

\author{Oren Barkan}
\authornote{Both authors contributed equally to this research.}
\affiliation{%
  \institution{The Open University}}
\author{Yehonatan Elisha}
\authornotemark[1]
\affiliation{%
  \institution{The Open University}}
\author{Jonathan Weill}
\affiliation{%
  \institution{Tel Aviv University}}
\author{Yuval Asher}
\affiliation{%
  \institution{Tel Aviv University}}
\author{Amit Eshel}
\affiliation{%
  \institution{Tel Aviv University}}
\author{Noam Koenigstein}
\affiliation{%
  \institution{Tel Aviv University}}

\begin{abstract}
This paper presents Deep Integrated Explanations (DIX) - a universal method for explaining vision models. DIX generates explanation maps by integrating information from the intermediate representations of the model, coupled with their corresponding gradients. Through an extensive array of both objective and subjective evaluations spanning diverse tasks, datasets, and model configurations, we showcase the efficacy of DIX in generating faithful and accurate explanation maps, while surpassing current state-of-the-art methods.
Our code is available at: \url{https://github.com/dix-cikm23/dix}
\end{abstract}

\maketitle

% \begin{CCSXML}

% \end{CCSXML}

% \ccsdesc[500]{Computing methodologies~Artificial intelligence}

% \keywords{Explainable AI, Deep Learning, Computer Vision}

 \section{Introduction}
\label{sec:intro}
The AI revolution has led to significant advancements across various application fields, including computer vision~\cite{He2016DeepRL, carion2020end_detr, dosovitskiy2020image_vit, lu2019vilbert, he2022masked_mae}, natural language processing~\cite{Mikolov:Word13,barkan2017bayesian,vaswani2017attention,devlin2018bert,barkan2020scalable,liu2019roberta,barkan2020bayesian,ginzburg2021self,barkan2021representation}, audio processing~\cite{defossez2018sing, engel2019gansynth, barkan2019deep, kumar2019melgan, barkan2019inversynth, barkan2023inversynth}, and recommender systems~\cite{he2017neural,wang2019neural,he2020lightgcn,malkiel2020recobert,barkan2019cb2cf,barkan2020attentive,barkan2020cold,barkan2021cold2, barkan2020neural, barkan2016item2vec,barkan2021anchor}.
Specifically, in computer vision, deep Convolutional Neural Networks (CNNs)~\cite{simonyan2014very, He2016DeepRL, Huang2017DenselyCC, Liu2022ACF}, alongside recent Vision Transformers (ViTs) models~\cite{dosovitskiy2020image} have risen to prominence, exhibiting outstanding performance in a variety of vision tasks~\cite{imagenet,he2017mask,carion2020end_detr,badrinarayanan2017segnet,Barkan_2023_discovery}.
This surge in popularity emphasizes the need to comprehend the underlying rationale driving the decisions and predictions of deep learning models.

Despite their remarkable achievements, most deep neural networks remain enigmatic, often considered black boxes due to their vast number of parameters and intricate non-linearities. This opacity has ignited the growth of explainable AI as a focal research area within the realm of deep learning. Consequently, numerous methodologies have been proposed for explaining the predictions of deep learning models in computer vision~\cite{zeiler2014visualizing,simonyan2013deep,selvaraju2017grad, chefer2021transformer,Barkan_2023_iia,barkan2023ltx,barkan2023six}, natural language processing~\cite{barkan2021grad,malkiel2022interpreting}, and recommender systems~\cite{barkan2020explainable,barkan2023modeling,gaiger2023not}.

Explanation techniques aim to bridge the gap in understanding by generating heatmap-like explanation maps. These maps spotlight distinct input regions, attributing predictions to specific areas within the input image. Initially, rooted in gradient-based approaches, early methods generated explanation maps by analyzing the gradient of predictions concerning the input image~\cite{simonyan2013deep,simonyan2014very,springenberg2014striving}. Subsequently, several works~\cite{selvaraju2017grad,chattopadhay2018grad,jiang2021layercam,barkan2021gam} proposed deriving explanation maps from the internal activation maps produced by the network, along with their gradients. Other techniques, such as Integrated Gradients (IG)~\cite{SundararajanTY17}, relying on path integration, created explanation maps by accumulating gradients from linear interpolations between input and reference images.

Predominantly applied to CNNs, the aforementioned methods arose before the emergence of Transformer-based architectures~\cite{vaswani2017attention}. With the advent of ViT models~\cite{dosovitskiy2020image}, a variety of methodologies were proposed to interpret and explain them, including recent explanation techniques like those presented in~\cite{chefer2021transformer,chefer2021generic}.

This paper introduces Deep Integrated Explanations (DIX), a comprehensive approach aimed at explaining vision models, which finds applicability across both CNN and ViT architectures. DIX employs integration over the internal model representations and their gradients, facilitating the extraction of insights from any activation (or attention) map within the network. 

We present a thorough objective and subjective evaluation, showcasing the efficacy of DIX on both CNN and ViT models. Our results reveal its superiority over other baselines across various explanation and segmentation tasks, encompassing diverse datasets, model architectures, and evaluation metrics. Additionally, we validate the credibility of DIX in producing faithful explanation maps through an extensive set of sanity tests, as outlined in~\cite{adebayo2018sanity}.

\section{Related Work}
\label{sec:related}
\subsection{Explanation Methods for CNNs} A diverse range of explanation methods were proposed for explaining CNN models, categorized into various types including perturbation-based methods, gradient methods, saliency-based methods,and gradient-free methods. 
Perturbation-based methods~\cite{fong2019understanding,fong2017interpretable} gauge output sensitivity concerning input through random perturbations applied in the input space.
Saliency-based methods~\cite{dabkowski2017real,simonyan2013deep,mahendran2016visualizing,zhou2016learning,zeiler2014visualizing,zhou2018interpreting} leverage feature maps obtained through forward propagation to interpret model predictions.

Gradient methods utilize prediction gradients with respect to the input or intermediate activation maps. These methods yield explanation maps based on the gradient itself or by a combination of the activation maps with their gradients~\cite{Shrikumar2016NotJA, Srinivas2019FullGradientRF}.
For instance, SmoothGrad~\cite{smilkov2017smoothgrad} presents a smoothing approach, applied by adding random Gaussian noise to the input image at each iteration. Another notable example is the Grad-CAM (GC)~\cite{selvaraju2017grad} method, which leverages activation maps from the final convolutional layer in conjunction with their pooled gradients to generate explanation maps. The effectiveness of GC has subsequently inspired numerous follow-up work~\cite{chattopadhay2018grad,Fu2020AxiombasedGT,jiang2021layercam,barkan2021gam}.

Gradient-free methods generate explanation maps by manipulating activation maps without relying on gradient information~\cite{Wang2020ScoreCAMSV,Desai2020AblationCAMVE}. For instance, LIFT-CAM~\cite{Jung2021TowardsBE} utilizes the DeepLIFT~\cite{Shrikumar2017LearningIF} technique to estimate SHAP values of activation maps~\cite{Lundberg2017AUA}, which are then combined with the activation maps to produce the explanation map. However, gradient-free methods have a drawback: they neglect gradient information, thereby constraining their ability to steer explanations toward the target or predicted class.

Finally, a notable avenue of research pertains to path integration methods. Integrated Gradients (IG)~\cite{SundararajanTY17} involves integration across interpolated image gradients. Blur IG (BIG)~\cite{xu2020attribution} focuses on introducing information using a baseline and adopts a path that gradually removes Gaussian blur from the attributed image. Guided IG (GIG)~\cite{kapishnikov2021guided} refines IG by introducing an adaptive path strategy. By computing integration along an alternative path, it circumvents high gradient regions, often resulting in a reduction of irrelevant attributions.

Distinguished from the aforementioned works, DIX employs integration, facilitating interpolation on the internal representations produced by the network, and offers to combine the resulting explanation maps from all network layers. Furthermore, DIX does not confine the integrand to simple gradients, but rather encompasses an arbitrary function involving the activation (attention) maps and their gradients.

% \vspace{-2mm}
\subsection{Explanation Methods for ViTs}
Early attempts to explain Transformers employed the attention scores inherent to ViT models in order to glean insights w.r.t. the input~\cite{vaswani2017attention,carion2020end}. 
However, it is not clear how to combine the scores from different layers. Simple averaging the attention scores of each token, for example, leads to blurring of the signal~\cite{chefer2021transformer}.

Abnar and Zuidema~\cite{abnar2020quantifying} proposed the Rollout method to compute attention scores to input tokens at each layer by considering raw attention scores in a layer as well as those from precedent layers. Rollout improved results over the utilization of a single attention layer.  However, by relying on simplistic aggregation assumptions, irrelevant tokens often become highlighted. LRP~\cite{bach2015pixel}, proposed to propagate gradients from the output layer to the beginning, considering all the components in the transformer's layers and not just the attention layers. 

Recently, Chefer et al.\cite{chefer2021transformer} introduced Transformer Attribution (T-Attr), a class-specific Deep Taylor Decomposition method that employs relevance propagation for both positive and negative attributions. More recently, the same authors introduced Generic Attention Explainability (GAE)\cite{chefer2021generic}, which is an extension of T-Attr aimed at explaining Bi-Modal transformers. T-Attr and GAE stand as state-of-the-art methods for explaining ViT models, exhibiting superior performance when compared to other effective explanation methods, including LRP and partial LRP~\cite{voita2019analyzing}.

DIX differs from T-Attr and GAE in two main aspects: First, DIX is a versatile method capable of producing explanation maps for both CNNs and ViTs. Second, in the context of ViT models, DIX employs path integration on the interpolated attention matrices while incorporating the Gradient Rollout (GR) representation (a variant of the Rollout method) as the function for integration.

\section{Deep Integrated Explanations}
\label{sec:method}
Let $f:\R^{D_0}\rightarrow \R^K$ be a neural network with $L$ hidden layers that takes an input (image) $\x \in \mathbb{R}^{D_0}$ and produces a prediction $f(\x)\in \R^K$. We denote $\x^l$ ($1\leq l\leq L$) as the intermediate representation generated by the $l$-th hidden layer in $f$ (based on the input $\x$), and $f^l:\R^{D_l}\rightarrow \R^K$ as the sub network of $f$ that takes $\x^l$ as an input and outputs the prediction $f(\x)$. Consequently, we have the relationship $f^l(\x^l)=f(\x)$.
Additionally, we denote $\x^0 = \x$ and $f^0 = f$.

Our assumption is that $\x^l$ preserves the spatial structure of $\x$ (though at a different resolution) such that each element in $x^l$ is associated with its corresponding elements in $\x$ (e.g., this assumption holds true for CNNs). W.l.o.g, we restrict the discussion to multi-class classification problems, hence $f$ outputs a vector assigning score to each class, and the score for the class $k$ is denoted as $f_k(\x)$. 

Our objective is to \emph{explain} the prediction $f_k(\x)$ for the class $k$. In this work, we define an \emph{explanation map} $\m^l$ as a tensor assigning an attribution score to each element in $\x^l$ w.r.t. the prediction $f^l_k(\x^l)=f_k(\x)$. Consequently, $\m^l$ must match the dimensions of $\x^l$. Note that our ultimate goal is to attribute the prediction to each element in the input $\x$, and due to the spatial structure preservation, each element in $\m^l$ can be associated with a set of elements in $\x$.

Let $\z^l \in \R^{D_l}$ be a baseline serving as a reference for the informative representation $\x^l$. $\z^l$ can be the null representation, random noise, or other baselines representing missing information. In what follows, we present a decomposition of the score difference $f_k(\x) - f^l_k(\z^l)$, from which an explanation map $\m^l$ is derived.

Let $C^l$ be a differentiable curve connecting $\z^l$ to $\x^l$. $C^l$ is parameterized by a vector function $\r^l:[0,1] \rightarrow \R^{D_l}$ such that $\r^l(0) = \z^l$ and $\r^l(1) = \x^l$. The score difference $f_k(\x) - f^l_k(\z^l)$ can then be expressed as follows:
\begin{equation}
\label{eq:decomp}
\begin{split}
    f_k(\x) - f^l_k(\z^l) &= f^l_k(\r^l(1)) - f^l_k(\r^l(0))
    \\&= \int_0^1 \frac{d}{dt} f^l_k(\r^l(t))dt
    \\&= \int_0^1 \nabla f^l_k(\r^l(t)) \cdot \frac{d\r^l(t)}{dt} dt
    \\&= \sum_{i=1}^{D_l} \int_0^1 g^l_i(t) h^l_i(t) dt,
\end{split}
\end{equation}
where
\begin{equation*}
g^l_i(t) = \frac{\partial f^l_k(\r^l(t))}{\partial r^l_i(t)} \quad \text{and} \quad h^l_i(t) = \frac{dr^l_i(t)}{dt},
\end{equation*}
with $\cdot$ representing the dot product operator, and $r^l_i(t)$ being the $i$-th element in the interpolant $\r^l(t)$. The first equality in Eq.~\ref{eq:decomp} consequents from the design of $\r^l$ and the fact that $f^l(\x^l)=f(\x)$. The second equality stems from the fundamental theorem of calculus. The third equality arises from the multivariate chain rule, and the last equality results from decomposing the dot product into a summation and then interchanging the order of finite sum and integration.

Equation~\ref{eq:decomp} breaks down the score difference into a sum, where each term is a line integral along the $i$-th element of curve $C^l$, and the integrand is a function involving the partial derivative of the prediction $f^l_k(\r^l(t))$ w.r.t. the $i$-th element in the interpolant $\r^l(t)$. Consequently, each term in the sum resembles the attribution of the prediction $f_k(\x)$ to an individual element in $\x^l$ through the integrated partial derivatives along $C^l$. Equipped with Eq.~\ref{eq:decomp}, an explanation map for $\x^l$ can be formed as follows:
\begin{equation}
\label{eq:m-l}
    \m^l= \int_0^1 \g^l(t) \circ \h^l(t)dt,
\end{equation}
where $\circ$ denotes the element-wise multiplication, $\g^l(t) = \frac{\partial f^l_k(\r^l(t))}{\partial \r^l(t)}$ is the gradient of the prediction w.r.t. the interpolant, and $\h^l(t) = \frac{d\r^l(t)}{dt}$. Note that 
$\m^l\in \R^{D_l}$ with $m^l_i=\int_0^1 g^l_i(t) h^l_i(t) dt$. Notable, for $l=0$, Eq.~\ref{eq:m-l} is equivalent to the IG~\cite{SundararajanTY17} explanation map, where the interpolation takes place in the input space.  

Equation~\ref{eq:m-l} integrates the gradients of the interpolated activation maps $\r^l(t)$. Empirically, we found that incorporating the information from $\r^l(t)$ itself (beyond its gradient) yields enhanced explanations, both visually and quantitatively. This observation is consistent with previous works~\cite{selvaraju2017grad, chattopadhay2018grad,barkan2021gam}. Furthermore, since for $l>0$, $\m^l$ does not match the spatial dimensions of the input $\x$, a subsequent transformation $\psi^l$ is employed to ensure a proper match. To this end, we define the DIX explanation map as follows: 
\begin{equation}
\label{eq:m-l-dix}
    \m^l_{\text{DIX}} = \psi^l\left(\int_0^1 \phi\left(\r^l(t), \g^l(t)\right) \circ \h^l(t)dt\right),
\end{equation}
where the exact implementation details of $\phi$ and $\psi$ are architecture dependent and are outlined in Sec.~\ref{sec:impl}.

In this work, we choose $C^l$ to be the linear curve connecting $\z^l$ to $\x^l$, hence
\begin{equation}
\label{eq:r-l}
   \r^l(t)=\z^l + t(\x^l - \z^l) \quad \text{and} \quad \h^l(t)=\x^l - \z^l.
\end{equation}
In practice, the integration in Eq.~\ref{eq:m-l-dix} is numerically approximated as follows:
\begin{equation}
\label{eq:m-l-dix-approx}
    \m^l_{\text{DIX}} \approx \psi^l\left(\frac{\x^l-\z^l}{N} \circ \sum_{n=1}^N \phi\left(\r^l \left(\frac{n}{N} \right),\g^l \left(\frac{n}{N} \right)\right)\right),
\end{equation}
where we have employed the linear interpolation from Eq.~\ref{eq:r-l}. In this work, we set $N=10$. The complexity of DIX is similar to IG, except for the extra computation induced by $\phi$ and $\psi^l$.

Given that different network layers capture varying types of information and resolution, we propose aggregating information from explanation maps produced for different values of $l$. As such, the final explanation map is constructed as follows:
\begin{equation}
\label{eq:m-dix-final}
   \m_{\text{DIX}}^S=\frac{1}{|S|} \sum_{l \in S} \m^l,
\end{equation}
where $S$ is a set indicating the layer indexes participating in the aggregation. Our experimentation indicates that the best-performing DIX configurations leverage a combination of explanation maps from the last two or three layers. Thus, in Sec.~\ref{subsec:results}, we report results for $S=\{L-1,L\}$ (\textbf{DIX2}) and $S=\{L-2,L-1,L\}$ (\textbf{DIX3}). However, for the sake of completeness, we also present results for $S=\{L\}$ (\textbf{DIX1}) as part of our ablation study in Sec.~\ref{sec:ablation-study}.

\subsection{Implementation Details}
\label{sec:impl}
In this section, we describe concrete implementations of DIX for both CNN and ViT architectures. 

\paragraph{CNN Models: }
In the case of CNNs, the architecture of $f$ consists of residual blocks~\cite{he2016deep} that produces 3D tensors representing the activation maps $\x^l$. Correspondingly, $\z^l$ is a 3D tensor where each channel is determined by broadcasting the minimum value of the respective activation map within $\x^l$. Furthermore, we set $\phi$ to the element-wise multiplication. 

We motivate this design choice by the fact that $\r^l(t)$ represents the interpolated activation map, highlighting regions where filters are activated and patterns are detected. Its gradient gauges the attribution degree of the specific class of interest to each element in the activation map. Thus, we expect that regions exhibiting both large gradient and activation (of the same sign) will yield effective explanations. This property is achieved through element-wise multiplication of $\r^l(t)$ by its gradient $\g^l(t)$. Finally, $\psi^l$ is set to the mean reduction on the channel axis followed by a resize operation yielding a 2D explanation map that matches the spatial dimensions of $\x$.

\paragraph{ViT Models: }
In ViT~\cite{dosovitskiy2020image_vit}, the architecture of $f$ consists of transformer encoder blocks producing 2D tensors (sequence of token representations). The input $\x$ is transformed to a 2D tensor as well, where the first token is the \textsc{[CLS]} token, and the rest of the tokens are representations of patches in the original image.

In our implementation, we choose to interpolate on the attention matrices, which in turn affect the output produced by the encoder block. Specifically, $\r^l(t)$ is a 3D tensor that accommodates all the attention matrices produced by the $l$-th encoder block. The reference $\z^l$ is set to the zero tensor (since the values in the attention matrix are in $[0,1]$). $\phi$ implements a variant of the Attention Rollout (AR) method~\cite{abnar2020quantifying} that we name Gradient Rollout (GR). GR is similar to AR, with a slight modification. Instead of operating solely on the plain attention matrices, GR initially performs an element-wise multiplication of the attention matrices by their corresponding gradients. Following this, GR proceeds with the original Rollout computation~\cite{abnar2020quantifying}, resulting in the first row of the derived matrix (associated with the \textsc{[CLS]} token). This output is further processed by truncating its initial element and reshaping it into a $14 \times 14$ matrix. The exact implementation of GR appears in our GitHub repository\footnote{It is worth noting that our experimental findings suggest comparable performance when substituting the matrix product operation with summation within the context of the GR computation}. Lastly, $\psi^l$ remains consistent across all layers, conducting a resize operation to align with the spatial dimensions of $\x$. 

\section{Experimental Setup}
\label{sec:experiments}
Our evaluation encompasses three distinct CNN architectures: ResNet101 (\textbf{RN})\cite{He2016DeepRL}, DenseNet201 (\textbf{DN})\cite{Huang2017DenselyCC}, and ConvNext-Base (\textbf{CN})\cite{Liu2022ACF}, and two different architectures of ViT: ViT-Base (\textbf{ViT-B}) and ViT-Small (\textbf{ViT-S})\cite{dosovitskiy2020image_vit}. The information regarding preprocessing methodologies and direct access to all the aforementioned models can be found in our GitHub repository. DIX is evaluated and compared to other explanation methods through a series of explanation, segmentation, and sanity tests.

\subsection{Explanation Metrics}
\label{subsec:explanation_metrics}
It is difficult to quantify the quality of explainability methods, and there is no single agreed-upon metric. The explanations metrics in this study aim to assess how well the explanations align with hypothetical changes (counterfactuals) to the input. Essentially, it's about asking ``what if'' questions regarding the input and determining whether the explanations provided are consistent with those hypothetical scenarios. To comprehensively evaluate our method, we carefully followed several prominent evaluation protocols.
\paragraph{Perturbation Tests} We followed the protocol from~\cite{chefer2021transformer}, which is the current state-of-the-art in explaining ViTs, and report the Negative Perturbation AUC (\textbf{NEG}) and the Positive Perturbation AUC (\textbf{POS}). NEG is a counterfactual test that entails a gradual blackout of the pixels in the original image in increasing order according to the explanation map while searching to see when the model's top predicted class changes. By masking pixels in increasing order,  we expect to remove the least relevant pixels first, and the model's top predicted class is expected to remain unchanged for as long as possible. Results are measured in terms of the Area Under the Curve (AUC), and higher values are considered better. Accordingly, the POS test entails masking the pixels in decreasing order with the expectation that the model's top predicted class will change quickly, hence in POS, lower values are better. 
In addition, we follow~\cite{petsiuk2018rise} and report the Insertion AUC (\textbf{INS}) and Deletion AUC (\textbf{DEL}) perturbation tests. 
INS and DEL entail a gradual blackout in increasing or decreasing order, similar to NEG and POS, respectively. 
However instead of tracking the point at which the top predicted class changes, in \textbf{INS} and DEL the AUC is computed with respect to the predicted probability of the top class. By masking pixels according to increasing/decreasing order of importance, we expect that the predicted probability of the top class will decrease slowly/quickly, respectively. Hence, for INS higher values are better and for DEL lower values are better.

\paragraph{ADP and PIC Tests}
We follow ~\cite{chattopadhay2018grad} and report the Average Drop Percentage (\textbf{ADP}) and the Percentage Increase in Confidence (\textbf{PIC}) tests. Both tests relate to the change in the probability of the predicted class after applying the mask to the original image. A good explanation map is expected to highlight the most significant regions for decision-making. Hence, applying such a mask can be seen as a removal of the ``background''. The ADP test measures the average percentage of model confidence drop after applying the mask. A good mask is expected to maintain the most relevant areas and minimize confidence drop, hence for ADP lower values are considered better. However, we note that ADP is a problematic metric since a naive all-ones mask yields an optimal ADP value of $0$. Nevertheless, we included it for the sake of compatibility with previous works~\cite{chattopadhay2018grad}. 
In some instances, the model's confidence increases after applying a good explanation mask that removes a confusing background. Hence, PIC is a binary test that measures the percentage of instances in which the model's confidence increased after applying the mask on the original input. For PIC higher values are considered better.

\paragraph{AIC and SIC Tests}

We follow ~\cite{kapishnikov2019xrai} and report the Accuracy Information Curve (\textbf{AIC}) and the Softmax Information Curve (\textbf{SIC}) tests. In these tests, we start with a completely blurred image and gradually sharpen the image areas that are deemed important by a given explanation method. Gradually sharpening the image areas increases the information content of the image. We then compare the explanation methods by measuring the approximate image entropy (e.g., compressed image size) and the model’s performance (e.g., model accuracy). 
The AIC metric measures the accuracy of a model as a function of the amount of information provided to the explanation method. AIC is defined as the AUC of the accuracy vs. information plot. The SIC metric measures the information content of the output of a softmax classifier as a function of the amount of information provided to the explanation method. SIC is defined as the AUC of the entropy vs. information plot. The entropy of the softmax output is a measure of the uncertainty or randomness of the classifier's predictions. For both AIC and SIC, the information provided to the method is quantified by the fraction of input features that are considered during the explanation process.

\subsection{Segmentation Metrics} 
\label{subsec:seg_metrics}
While possessing a superior segmentation capability does not necessarily imply a superior explanatory aptitude, we undertake this evaluation task for the sake of completeness in our comparison with previous works assessing this aspect \cite{chefer2021transformer,chefer2021generic,jiang2021layercam,wang2020score}. Segmentation accuracy is assessed according to the following metrics: Pixel Accuracy (\textbf{PA}), mean-intersection-over-union (\textbf{mIoU}), mean-average-precision (\textbf{mAP}), and the mean-F1 score (\textbf{mF1})~\cite{chefer2021transformer}. 

% \vspace{-3mm}
\subsection{Datasets}
Explanation maps are produced for the ImageNet~\cite{imagenet} ILSVRC 2012 (\textbf{IN}) validation set, consisting of 50K images from 1000 classes. We follow the same setup from~\cite{chefer2021transformer}, where for each image, an explanation map is produced w.r.t. the class predicted by the model.
% For segmentation tests are conducted on three datasets:
Segmentation tests are conducted on three datasets:
(1) ImageNet-Segmentation~\cite{guillaumin2014imagenet} (\textbf{IN-Seg}): This is a subset of ImageNet validation set consisting of 4,276 images from 445 classes for which annotated segmentations are available. (2) Microsoft Common Objects in COntext 2017~\cite{lin2014microsoft} (\textbf{COCO}): This is a validation set that contains 5,000 annotated segmentation images from 80 different classes. Some images consist of multi-label annotations (multiple annotated objects). In our evaluation, all annotated objects in the image are considered as the ground-truth. (3) PASCAL Visual Object Classes  2012~\cite{Everingham2009ThePV} (\textbf{VOC}): This is a validation set that contains annotated segmentations for 1,449 images from 20 classes.

\subsection{Evaluated Methods}
Our evaluation encompasses a comprehensive assessment of various explanation methods, including gradient-based approaches, path-integration techniques, as well as gradient-free methods. 

For CNN models, the following explanation techniques are considered: 
Integrated Gradients (\textbf{IG})~\cite{SundararajanTY17}, Guided IG (\textbf{GIG})~\cite{kapishnikov2021guided}, Blur IG (\textbf{BIG})~\cite{xu2020attribution}, Ablation-CAM (\textbf{AC})~\cite{Desai2020AblationCAMVE}, Layer-CAM (\textbf{LC})~\cite{jiang2021layercam}, LIFT-CAM (\textbf{LIFT})~\cite{Jung2021TowardsBE},  
Grad-CAM (\textbf{GC})~\cite{selvaraju2017grad}, Grad-CAM++ (\textbf{GC++})~\cite{chattopadhay2018grad}, X-Grad-CAM (\textbf{XGC})~\cite{Fu2020AxiombasedGT}, and FullGrad (\textbf{FG})~\cite{Srinivas2019FullGradientRF}.

For ViT models, we consider two state-of-the-art methods: Transformer Attribution (\textbf{T-Attr})~\cite{chefer2021transformer} and Generic Attention Explainability (\textbf{GAE})~\cite{chefer2021generic}. Both methods were shown to outperform other strong baselines such as partial LRP~\cite{voita2019analyzing}, and GC~\cite{chefer2021generic} for transformers. A detailed description of all explanation methods is provided in our GitHub repository.
Lastly, our universal DIX method is evaluated on both CNNs and ViTs, where we consider two versions: DIX2 and DIX3 following the description in Sec.~\ref{sec:method}.

\subsection{Sanity Tests for Explanation Methods}
\label{subsec: santests_description}
To comprehensively assess the robustness and credibility of DIX, we conducted the \emph{parameter randomization} and \emph{data randomization} sanity tests as outlined in~\cite{adebayo2018sanity}. For these evaluations, we employed DIX3, along with the VGG-19\cite{vgg} model and the IN dataset.

\paragraph{Parameter Randomization Test}
% \label{subsubsec:modelrand_test}
This test compares explanation maps generated by the explanation method under two model setups: (1) Trained - a model trained on the dataset (e.g., pretrained VGG-19 on ImageNet), and (2) Random - the same model architecture with randomized weights. Significant differences in explanation maps between the trained and random models suggest sensitivity to model parameters. Conversely, similar maps indicate insensitivity and limited utility for model explanation. 

Two types of parameter randomization tests are conducted on a trained model:

(1) \emph{Cascading Randomization}: We randomize model weights layer by layer, starting from the top and progressing to the bottom. This process randomizes the learned weights from top to bottom.

(2) \emph{Independent Randomization}: We randomize each individual layer's weights, one layer at a time (while keeping all other layers' weights fixed). This allows us to evaluate each layer's impact on explanation map sensitivity, independently.

In both tests, we compare the resulting explanations obtained by using the model with random weights to those derived from the original weights of the model.

\paragraph{Data Randomization Test}
The data randomization test assesses an explanation method's sensitivity to data labeling. We compare explanation maps for two models with identical architectures, but trained on different datasets: one with original labels and another with labels randomly permuted. Sensitivity to labeling is indicated by significantly different explanation maps, while insensitivity suggests independence from instance-label relationships. To conduct the data randomization test, we permute the training labels in the dataset and train the model to achieve a training set accuracy greater than 95\%. Note that the resulting model's test accuracy is never better than randomly guessing a label. We then compute explanations on the same test inputs for both the model trained on true labels and the model trained on randomly permuted labels.

\begin{table*}[t]
  \caption{Explanation tests results on the IN dataset (CNN models): For POS, DEL and ADP, lower is better. For NEG, INS, PIC, SIC and AIC, higher is better. See Sec.~\ref{subsubsection:obj2} for details.}
  \label{tab:cnn_backbones_metrics_exp}
\begin{center}
   \scalebox{1.02}{
    \begin{tabular}
    {@{}lc@{}lc@{}lc@{}lc@{}lc@{}lc@{}lc@{}lc@{}}
    \toprule
      & & \multicolumn{1}{l}{GC} & \multicolumn{1}{l}{GC++} & \multicolumn{1}{l}{LIFT} & \multicolumn{1}{l}{AC} & \multicolumn{1}{l}{IG}& \multicolumn{1}{l}{GIG}& \multicolumn{1}{l}{BIG} & \multicolumn{1}{l}{FG} & \multicolumn{1}{l}{LC} & \multicolumn{1}{l}{XGC} & \multicolumn{1}{l}{DIX2} & \multicolumn{1}{l}{DIX3}\\
    \midrule
    \multirow{8}{*}{RN}

    & \multirow{1}{*}{NEG} & \multicolumn{1}{l}{\underline{56.41}} & \multicolumn{1}{l}{55.20} & \multicolumn{1}{l}{55.39} & \multicolumn{1}{l}{54.98} & \multicolumn{1}{l}{45.66} &\multicolumn{1}{l}{43.97} & \multicolumn{1}{l}{42.25}& \multicolumn{1}{l}{54.81} & \multicolumn{1}{l}{53.52} & \multicolumn{1}{l}{53.46} & \multicolumn{1}{l}{56.28} & \multicolumn{1}{l}{\textbf{57.13}} \\

    & \multirow {1}{*}{POS}  & \multicolumn{1}{l}{17.82} & \multicolumn{1}{l}{18.01} & \multicolumn{1}{l}{17.53} & \multicolumn{1}{l}{19.38} & \multicolumn{1}{l}{17.24} &\multicolumn{1}{l}{17.68} & \multicolumn{1}{l}{17.44} & \multicolumn{1}{l}{18.06} & \multicolumn{1}{l}{17.92} & \multicolumn{1}{l}{21.02} & \multicolumn{1}{l}{\textbf{15.69}} & \multicolumn{1}{l}{\underline{17.11}}\\

    & \multirow {1}{*}{INS}   & \multicolumn{1}{l}{\underline{48.14}} & \multicolumn{1}{l}{47.56} & \multicolumn{1}{l}{45.39} & \multicolumn{1}{l}{47.05} & \multicolumn{1}{l}{39.87}  &\multicolumn{1}{l}{37.92} & \multicolumn{1}{l}{36.04}& \multicolumn{1}{l}{42.68} & \multicolumn{1}{l}{46.11} & \multicolumn{1}{l}{43.26} & \multicolumn{1}{l}{48.09} & \multicolumn{1}{l}{\textbf{48.91}}\\

    & \multirow {1}{*}{DEL}   & \multicolumn{1}{l}{13.97} & \multicolumn{1}{l}{14.17} & \multicolumn{1}{l}{15.32} & \multicolumn{1}{l}{14.23} & \multicolumn{1}{l}{13.49} &\multicolumn{1}{l}{14.18} & \multicolumn{1}{l}{13.95}& \multicolumn{1}{l}{14.64} & \multicolumn{1}{l}{14.31} & \multicolumn{1}{l}{14.98} & \multicolumn{1}{l}{\textbf{12.84}} & \multicolumn{1}{l}{\underline{13.36}}\\

     & \multirow {1}{*}{ADP}   & \multicolumn{1}{l}{17.87} & \multicolumn{1}{l}{16.91} & \multicolumn{1}{l}{18.03} & \multicolumn{1}{l}{16.18} & \multicolumn{1}{l}{37.52} &\multicolumn{1}{l}{35.28} & \multicolumn{1}{l}{40.85}& \multicolumn{1}{l}{21.06} & \multicolumn{1}{l}{24.34} & \multicolumn{1}{l}{17.02} & \multicolumn{1}{l}{\textbf{15.68}} & \multicolumn{1}{l}{\underline{16.02}}\\

    & \multirow {1}{*}{PIC} &  \multicolumn{1}{l}{36.69} & \multicolumn{1}{l}{36.53} & \multicolumn{1}{l}{35.95} & \multicolumn{1}{l}{35.52} & \multicolumn{1}{l}{19.94} &\multicolumn{1}{l}{18.72} & \multicolumn{1}{l}{24.53}& \multicolumn{1}{l}{31.59} & \multicolumn{1}{l}{35.43} & \multicolumn{1}{l}{36.18} & \multicolumn{1}{l}{\textbf{40.21}} & \multicolumn{1}{l}{\underline{37.29}}\\

    & \multirow {1}{*}{SIC}   &  \multicolumn{1}{l}{76.91} & \multicolumn{1}{l}{76.44} & \multicolumn{1}{l}{76.73} & \multicolumn{1}{l}{73.36} & \multicolumn{1}{l}{54.67} & \multicolumn{1}{l}{55.04} & \multicolumn{1}{l}{56.98} & \multicolumn{1}{l}{75.35} & \multicolumn{1}{l}{73.93} & \multicolumn{1}{l}{72.64} & \multicolumn{1}{l}{\underline{77.61}} & \multicolumn{1}{l}{\textbf{78.12}}\\

        & \multirow {1}{*}{AIC}   &  \multicolumn{1}{l}{74.36} & \multicolumn{1}{l}{71.97} & \multicolumn{1}{l}{72.76} & \multicolumn{1}{l}{70.35} & \multicolumn{1}{l}{51.92} & \multicolumn{1}{l}{53.38} & \multicolumn{1}{l}{53.36} & \multicolumn{1}{l}{71.49} & \multicolumn{1}{l}{65.77} & \multicolumn{1}{l}{69.85} & \multicolumn{1}{l}{\underline{76.09}} & \multicolumn{1}{l}{\textbf{76.34}}\\

    \midrule
    \multirow{8}{*}{CN}

    & \multirow {1}{*}{NEG}   & \multicolumn{1}{l}{52.86} & \multicolumn{1}{l}{53.82} & \multicolumn{1}{l}{53.98} & \multicolumn{1}{l}{53.68} & \multicolumn{1}{l}{45.24} &\multicolumn{1}{l}{41.43} & \multicolumn{1}{l}{40.72}& \multicolumn{1}{l}{52.06} & \multicolumn{1}{l}{54.12} & \multicolumn{1}{l}{52.13} & \multicolumn{1}{l}{\underline{54.40}} & \multicolumn{1}{l}{\textbf{55.23}}\\

    & \multirow {1}{*}{POS}   &  \multicolumn{1}{l}{17.52} & \multicolumn{1}{l}{17.85} & \multicolumn{1}{l}{18.23} & \multicolumn{1}{l}{18.19} & \multicolumn{1}{l}{17.42} &\multicolumn{1}{l}{18.03} & \multicolumn{1}{l}{18.14}& \multicolumn{1}{l}{18.26} & \multicolumn{1}{l}{17.58} & \multicolumn{1}{l}{20.83} & \multicolumn{1}{l}{\underline{16.96}} & \multicolumn{1}{l}{\textbf{16.51}}\\

    & \multirow {1}{*}{INS}   &  \multicolumn{1}{l}{45.65} & \multicolumn{1}{l}{45.19} & \multicolumn{1}{l}{43.86} & \multicolumn{1}{l}{49.18} & \multicolumn{1}{l}{37.22} 
    &\multicolumn{1}{l}{32.99} & \multicolumn{1}{l}{31.02}& \multicolumn{1}{l}{42.01} & \multicolumn{1}{l}{44.14} & \multicolumn{1}{l}{42.07} & \multicolumn{1}{l}{\underline{49.53}} & \multicolumn{1}{l}{\textbf{49.86}}\\

         & \multirow {1}{*} {DEL} & \multicolumn{1}{l}{13.43} & \multicolumn{1}{l}{14.17} & \multicolumn{1}{l}{15.18} & \multicolumn{1}{l}{14.73} & \multicolumn{1}{l}{12.36} 
         &\multicolumn{1}{l}{13.08} & \multicolumn{1}{l}{13.29}& \multicolumn{1}{l}{14.21} & \multicolumn{1}{l}{13.64} & \multicolumn{1}{l}{14.78} & \multicolumn{1}{l}{\underline{11.95}} & \multicolumn{1}{l}{\textbf{11.74}}\\

         & \multirow {1}{*}{ADP}   & \multicolumn{1}{l}{22.46} & \multicolumn{1}{l}{22.35} & \multicolumn{1}{l}{29.13} & \multicolumn{1}{l}{24.38} & \multicolumn{1}{l}{36.98} 
         &\multicolumn{1}{l}{35.79} & \multicolumn{1}{l}{41.73}& \multicolumn{1}{l}{30.75} & \multicolumn{1}{l}{37.62} & \multicolumn{1}{l}{25.68} & \multicolumn{1}{l}{\underline{22.24}} & \multicolumn{1}{l}{\textbf{22.19}}\\

        & \multirow {1}{*}{PIC}   &    \multicolumn{1}{l}{23.16} & \multicolumn{1}{l}{24.42} & \multicolumn{1}{l}{22.34} & \multicolumn{1}{l}{24.59} & \multicolumn{1}{l}{17.65} 
        &\multicolumn{1}{l}{13.12} & \multicolumn{1}{l}{20.69}& \multicolumn{1}{l}{22.13} & \multicolumn{1}{l}{22.17} & \multicolumn{1}{l}{23.26} & \multicolumn{1}{l}{\underline{28.31}} & \multicolumn{1}{l}{\textbf{28.47}}\\

        & \multirow {1}{*}{SIC}   &  \multicolumn{1}{l}{65.93} & \multicolumn{1}{l}{67.94} & \multicolumn{1}{l}{54.75} & \multicolumn{1}{l}{63.95} & \multicolumn{1}{l}{53.36} & \multicolumn{1}{l}{58.35} & \multicolumn{1}{l}{57.27} & \multicolumn{1}{l}{62.84} & \multicolumn{1}{l}{69.11} & \multicolumn{1}{l}{59.12} & \multicolumn{1}{l}{\underline{69.83}} & \multicolumn{1}{l}{\textbf{70.18}}\\

        & \multirow {1}{*}{AIC}   &  \multicolumn{1}{l}{75.64} & \multicolumn{1}{l}{75.52} & \multicolumn{1}{l}{57.06} & \multicolumn{1}{l}{71.53} & \multicolumn{1}{l}{51.68} & \multicolumn{1}{l}{55.82} & \multicolumn{1}{l}{53.82} & \multicolumn{1}{l}{67.15} & \multicolumn{1}{l}{75.41} & \multicolumn{1}{l}{62.38} & \multicolumn{1}{l}{\underline{76.44}} & \multicolumn{1}{l}{\textbf{77.29}}\\

\midrule
    \multirow{8}{*}{DN}

    & \multirow {1}{*}{NEG}   & \multicolumn{1}{l}{\underline{57.40}} & \multicolumn{1}{l}{57.16} & \multicolumn{1}{l}{58.01} & \multicolumn{1}{l}{56.63} & \multicolumn{1}{l}{40.74} 
    &\multicolumn{1}{l}{37.31} & \multicolumn{1}{l}{36.67}& \multicolumn{1}{l}{56.79} & \multicolumn{1}{l}{56.96} & \multicolumn{1}{l}{55.74} & \multicolumn{1}{l}{57.31} & \multicolumn{1}{l}{\textbf{58.25}}\\

    & \multirow {1}{*}{POS}   &  \multicolumn{1}{l}{17.75} & \multicolumn{1}{l}{17.81} & \multicolumn{1}{l}{18.87} & \multicolumn{1}{l}{18.67} & \multicolumn{1}{l}{17.31} 
    &\multicolumn{1}{l}{17.46} & \multicolumn{1}{l}{17.38}& \multicolumn{1}{l}{17.84} & \multicolumn{1}{l}{17.62} & \multicolumn{1}{l}{18.67} & \multicolumn{1}{l}{\textbf{16.59}} & \multicolumn{1}{l}{\underline{17.14}} \\

    & \multirow {1}{*}{INS}   & \multicolumn{1}{l}{\underline{51.09}} & \multicolumn{1}{l}{50.89} & \multicolumn{1}{l}{50.63} & \multicolumn{1}{l}{50.41} & \multicolumn{1}{l}{37.58} 
    &\multicolumn{1}{l}{33.31} & \multicolumn{1}{l}{31.32}& \multicolumn{1}{l}{50.44} & \multicolumn{1}{l}{50.60} & \multicolumn{1}{l}{49.62} & \multicolumn{1}{l}{50.97} & \multicolumn{1}{l}{\textbf{51.58}}\\

         & \multirow {1}{*}{DEL}   & \multicolumn{1}{l}{13.61} & \multicolumn{1}{l}{13.63} & \multicolumn{1}{l}{13.29} & \multicolumn{1}{l}{15.31} & \multicolumn{1}{l}{13.26} 
         &\multicolumn{1}{l}{13.27} & \multicolumn{1}{l}{13.54}& \multicolumn{1}{l}{14.34} & \multicolumn{1}{l}{13.85} & \multicolumn{1}{l}{14.75} & \multicolumn{1}{l}{\textbf{12.73}} & \multicolumn{1}{l}{\underline{12.98}}\\

         & \multirow {1}{*}{ADP}   & \multicolumn{1}{l}{17.46} & \multicolumn{1}{l}{17.01} & \multicolumn{1}{l}{19.45} & \multicolumn{1}{l}{17.13} & \multicolumn{1}{l}{35.61}
         &\multicolumn{1}{l}{34.51} & \multicolumn{1}{l}{40.04}& \multicolumn{1}{l}{20.21} & \multicolumn{1}{l}{24.23} & \multicolumn{1}{l}{19.59} & \multicolumn{1}{l}{\textbf{16.29}} & \multicolumn{1}{l}{\underline{16.58}}\\

        & \multirow {1}{*}{PIC}   &  \multicolumn{1}{l}{34.68} & \multicolumn{1}{l}{35.21} & \multicolumn{1}{l}{34.13} & \multicolumn{1}{l}{31.22} & \multicolumn{1}{l}{22.35} 
        &\multicolumn{1}{l}{16.62} & \multicolumn{1}{l}{26.18}& \multicolumn{1}{l}{31.05} & \multicolumn{1}{l}{33.81} & \multicolumn{1}{l}{30.39} & \multicolumn{1}{l}{\textbf{38.91}} & \multicolumn{1}{l}{\underline{37.78}}\\

        & \multirow {1}{*}{SIC}   &  \multicolumn{1}{l}{75.62} & \multicolumn{1}{l}{74.75} & \multicolumn{1}{l}{74.72} & \multicolumn{1}{l}{73.94} & \multicolumn{1}{l}{54.59} & \multicolumn{1}{l}{58.55} & \multicolumn{1}{l}{57.66} & \multicolumn{1}{l}{72.93} & \multicolumn{1}{l}{74.34} & \multicolumn{1}{l}{73.94} & \multicolumn{1}{l}{\underline{77.24}} & \multicolumn{1}{l}{\textbf{77.32}}\\

        & \multirow {1}{*}{AIC}   &  \multicolumn{1}{l}{74.22} & \multicolumn{1}{l}{71.82} & \multicolumn{1}{l}{72.65} & \multicolumn{1}{l}{70.21} & \multicolumn{1}{l}{54.74} & \multicolumn{1}{l}{54.56} & \multicolumn{1}{l}{56.08} & \multicolumn{1}{l}{70.63} & \multicolumn{1}{l}{71.82} & \multicolumn{1}{l}{70.12} & \multicolumn{1}{l}{\underline{75.98}} & \multicolumn{1}{l}{\textbf{76.39}}\\

        \bottomrule
      \end{tabular}}
  \end{center}
    \end{table*}
\begin{table}

\caption{Explanation tests results on the IN dataset (ViT models): For POS, DEL and ADP, lower is better. For NEG, INS, PIC, SIC and AIC, higher is better. See Sec.~\ref{subsubsection:obj2} for details.
  }
  \label{tab:appendix_vit_backbones_metrics}  
  \begin{center}

  \scalebox{1}{
    \begin{tabular}{@{}lc@{}lc@{}lc@{}lc@{}}
    \toprule
      & & \multicolumn{1}{l}{T-Attr} & \multicolumn{1}{l}{GAE} & \multicolumn{1}{l}{DIX2} & \multicolumn{1}{l}{DIX3} \\
    \midrule
    \multirow{8}{*}{ViT-B}
    
        & \multirow {1}{*}{NEG} & \multicolumn{1}{l}{{54.16}} & \multicolumn{1}{l}{{54.61}} & \multicolumn{1}{l}{\underline{56.43}} & \multicolumn{1}{l}{\textbf{56.94}}\\

        & \multirow {1}{*}{POS} &  \multicolumn{1}{l}{17.03} & \multicolumn{1}{l}{17.32} & \multicolumn{1}{l}{\underline{15.10}} & \multicolumn{1}{l}{\textbf{14.85}}\\

        & \multirow {1}{*}{INS} & \multicolumn{1}{l}{48.58} & \multicolumn{1}{l}{48.96} & \multicolumn{1}{l}{\underline{49.51}} & \multicolumn{1}{l}{\textbf{50.59}}\\

        & \multirow {1}{*}{DEL} & \multicolumn{1}{l}{{14.20}} & \multicolumn{1}{l}{{14.37}} & \multicolumn{1}{l}{\underline{12.62}} & \multicolumn{1}{l}{\textbf{12.16}}\\
           
         & \multirow {1}{*}{ADP} & \multicolumn{1}{l}{{54.02}} & \multicolumn{1}{l}{{37.84}} & \multicolumn{1}{l}{\underline{35.93}} & \multicolumn{1}{l}{\textbf{35.58}}\\

        & \multirow {1}{*}{PIC} & \multicolumn{1}{l}{13.37} & \multicolumn{1}{l}{23.65}  & \multicolumn{1}{l}{\textbf{28.21}} & \multicolumn{1}{l}{\underline{27.41}}\\

        & \multirow {1}{*}{SIC} & \multicolumn{1}{l}{68.59} & \multicolumn{1}{l}{68.35}  & \multicolumn{1}{l}{\underline{68.94}} & \multicolumn{1}{l}{\textbf{69.11}}\\

        & \multirow {1}{*}{AIC} & \multicolumn{1}{l}{61.34} & \multicolumn{1}{l}{57.92}  & \multicolumn{1}{l}{\underline{62.42}} & \multicolumn{1}{l}{\textbf{65.03}}\\
             
        \midrule
        
    \multirow{8}{*}{ViT-S}
    
        & \multirow {1}{*}{NEG} & \multicolumn{1}{l}{53.29} & \multicolumn{1}{l}{52.81}  & \multicolumn{1}{l}{\underline{55.98}} & \multicolumn{1}{l}{\textbf{56.13}}\\

        & \multirow {1}{*}{POS} & \multicolumn{1}{l}{14.16} & \multicolumn{1}{l}{14.75}  & \multicolumn{1}{l}{\underline{13.09}} & \multicolumn{1}{l}{\textbf{12.32}}\\

        & \multirow {1}{*}{INS} & \multicolumn{1}{l}{45.72} & \multicolumn{1}{l}{45.21}  & \multicolumn{1}{l}{\underline{46.62}} & \multicolumn{1}{l}{\textbf{47.36}}\\

         & \multirow {1}{*}{DEL} & \multicolumn{1}{l}{11.28} & \multicolumn{1}{l}{11.92}  & \multicolumn{1}{l}{\underline{11.18}} & \multicolumn{1}{l}{\textbf{10.56}}\\

         & \multirow {1}{*}{ADP} & \multicolumn{1}{l}{51.94} & \multicolumn{1}{l}{36.98} & \multicolumn{1}{l}{\textbf{36.31}} & \multicolumn{1}{l}{\underline{36.57}}\\

        & \multirow {1}{*}{PIC} &   \multicolumn{1}{l}{13.67} &  \multicolumn{1}{l}{8.68} & \multicolumn{1}{l}{\textbf{18.39}} & \multicolumn{1}{l}{\underline{18.25}}\\

        & \multirow {1}{*}{SIC} &  \multicolumn{1}{l}{69.46} & \multicolumn{1}{l}{70.19}  & \multicolumn{1}{l}{\underline{70.92}} & \multicolumn{1}{l}{\textbf{71.55}}\\

        & \multirow {1}{*}{AIC} &  \multicolumn{1}{l}{63.86} & \multicolumn{1}{l}{64.49}  & \multicolumn{1}{l}{\underline{65.17}} & \multicolumn{1}{l}{\textbf{65.58}}\\
            
        \bottomrule
  \end{tabular}}

  \end{center}
  \vspace{-2mm}
\end{table}
\begin{table}

    \caption{Segmentation tests on three datasets (CNN models). For all metrics, higher is better. See Sec.~\ref{subsec:res_seg} for details.}
  \label{tab:cnns_segmentation_table}
   \begin{center}

   \scalebox{1}{
    \begin{tabular}{@{}lc@{}lc@{}lc@{}lc@{}lc@{}lc@{}lc@{}}
    \toprule
      & & & \multicolumn{1}{l}{GC} & \multicolumn{1}{l}{GC++} & \multicolumn{1}{l}{LIFT} & \multicolumn{1}{l}{AC} & \multicolumn{1}{l}{DIX2} & \multicolumn{1}{l}{DIX3}\\
    \midrule
    \multirow{12}{*}{IN-SEG}
    & \multirow{4}{*}{CN}

    & \multicolumn{1}{l}{PA} & \multicolumn{1}{l}{77.01} & \multicolumn{1}{l}{77.54} & \multicolumn{1}{l}{63.77} & \multicolumn{1}{l}{77.04} & \multicolumn{1}{l}{\underline{78.32}} & \multicolumn{1}{l}{\textbf{78.93}}\\

    & & \multicolumn{1}{l}{mAP} & \multicolumn{1}{l}{81.01} & \multicolumn{1}{l}{85.63} & \multicolumn{1}{l}{69.40} & \multicolumn{1}{l}{86.93} & \multicolumn{1}{l}{\underline{87.13}} & \multicolumn{1}{l}{\textbf{87.34}}\\

    & & \multicolumn{1}{l}{mIoU} & \multicolumn{1}{l}{56.58} & \multicolumn{1}{l}{58.35} & \multicolumn{1}{l}{53.81} & \multicolumn{1}{l}{58.42} & \multicolumn{1}{l}{\underline{58.64}} & \multicolumn{1}{l}{\textbf{58.79}}\\
    
    & & \multicolumn{1}{l}{mF1} & \multicolumn{1}{l}{36.88} & \multicolumn{1}{l}{38.26} & \multicolumn{1}{l}{35.91} & \multicolumn{1}{l}{41.29} & \multicolumn{1}{l}{\underline{42.51}} & \multicolumn{1}{l}{\textbf{42.95}}\\
    
    & \multirow{4}{*}{RN}

    & \multicolumn{1}{l}{PA} & \multicolumn{1}{l}{71.93} & \multicolumn{1}{l}{71.96} & \multicolumn{1}{l}{71.68} & \multicolumn{1}{l}{70.36} & \multicolumn{1}{l}{\underline{72.43}} & \multicolumn{1}{l}{\textbf{73.17}}\\

    & & \multicolumn{1}{l}{mAP} & \multicolumn{1}{l}{84.21} & \multicolumn{1}{l}{84.23} & \multicolumn{1}{l}{83.79} & \multicolumn{1}{l}{81.14} & \multicolumn{1}{l}{\underline{84.58}} & \multicolumn{1}{l}{\textbf{85.37}}\\

    & & \multicolumn{1}{l}{mIoU} & \multicolumn{1}{l}{53.06} & \multicolumn{1}{l}{53.29} & \multicolumn{1}{l}{52.17} & \multicolumn{1}{l}{52.91} & \multicolumn{1}{l}{\underline{53.93}} & \multicolumn{1}{l}{\textbf{54.16}}\\
    
    & & \multicolumn{1}{l}{mF1} & \multicolumn{1}{l}{42.51} & \multicolumn{1}{l}{42.68} & \multicolumn{1}{l}{41.95} & \multicolumn{1}{l}{42.08} & \multicolumn{1}{l}{\underline{42.75}} & \multicolumn{1}{l}{\textbf{43.18}}\\

    & \multirow{4}{*}{DN}

    & \multicolumn{1}{l}{PA} & \multicolumn{1}{l}{73.00} & \multicolumn{1}{l}{73.21} & \multicolumn{1}{l}{72.87} & \multicolumn{1}{l}{72.44} & \multicolumn{1}{l}{\underline{73.58}} & \multicolumn{1}{l}{\textbf{73.90}}\\

    & & \multicolumn{1}{l}{mAP} & \multicolumn{1}{l}{85.04} & \multicolumn{1}{l}{85.53} & \multicolumn{1}{l}{84.82} & \multicolumn{1}{l}{84.62} & \multicolumn{1}{l}{\underline{85.57}} & \multicolumn{1}{l}{\textbf{85.98}}\\

    & & \multicolumn{1}{l}{mIoU} & \multicolumn{1}{l}{54.18} & \multicolumn{1}{l}{54.57} & \multicolumn{1}{l}{54.11} & \multicolumn{1}{l}{54.89} & \multicolumn{1}{l}{\underline{55.42}} & \multicolumn{1}{l}{\textbf{56.03}}\\
    
    & & \multicolumn{1}{l}{mF1} & \multicolumn{1}{l}{41.74} & \multicolumn{1}{l}{42.58} & \multicolumn{1}{l}{41.61} & \multicolumn{1}{l}{43.51} & \multicolumn{1}{l}{\underline{43.71}} & \multicolumn{1}{l}{\textbf{43.79}}\\
    
    \midrule
    
    \multirow{12}{*}{COCO}

    & \multirow{4}{*}{CN}
    & \multicolumn{1}{l}{PA} & \multicolumn{1}{l}{68.75} & \multicolumn{1}{l}{66.49} & \multicolumn{1}{l}{60.37} & \multicolumn{1}{l}{64.10} & \multicolumn{1}{l}{\underline{68.87}} & \multicolumn{1}{l}{\textbf{69.38}}\\

    & & \multicolumn{1}{l}{mAP} & \multicolumn{1}{l}{75.02} & \multicolumn{1}{l}{75.21} & \multicolumn{1}{l}{67.98} & \multicolumn{1}{l}{76.09} & \multicolumn{1}{l}{\underline{76.94}} & \multicolumn{1}{l}{\textbf{77.43}}\\

    & & \multicolumn{1}{l}{mIoU} & \multicolumn{1}{l}{43.46} & \multicolumn{1}{l}{44.01} & \multicolumn{1}{l}{37.08} & \multicolumn{1}{l}{44.27} & \multicolumn{1}{l}{\underline{44.89}} & \multicolumn{1}{l}{\textbf{45.06}}\\
    
    & & \multicolumn{1}{l}{mF1} & \multicolumn{1}{l}{28.96} & \multicolumn{1}{l}{29.85} & \multicolumn{1}{l}{26.92} & \multicolumn{1}{l}{30.81} & \multicolumn{1}{l}{\underline{31.28}} & \multicolumn{1}{l}{\textbf{31.99}}\\
    
    & \multirow{4}{*}{RN}

    & \multicolumn{1}{l}{PA} & \multicolumn{1}{l}{64.17} & \multicolumn{1}{l}{64.39} & \multicolumn{1}{l}{64.02} & \multicolumn{1}{l}{63.90} & \multicolumn{1}{l}{\underline{64.75}} & \multicolumn{1}{l}{\textbf{64.94}}\\

    & & \multicolumn{1}{l}{mAP} & \multicolumn{1}{l}{74.19} & \multicolumn{1}{l}{74.27} & \multicolumn{1}{l}{73.78} & \multicolumn{1}{l}{72.80} & \multicolumn{1}{l}{\underline{74.38}} & \multicolumn{1}{l}{\textbf{74.91}}\\

    & & \multicolumn{1}{l}{mIoU} & \multicolumn{1}{l}{42.37} & \multicolumn{1}{l}{43.25} & \multicolumn{1}{l}{42.59} & \multicolumn{1}{l}{42.88} & \multicolumn{1}{l}{\underline{43.54}} & \multicolumn{1}{l}{\textbf{43.87}}\\
    
    & & \multicolumn{1}{l}{mF1} & \multicolumn{1}{l}{31.64} & \multicolumn{1}{l}{32.82} & \multicolumn{1}{l}{31.77} & \multicolumn{1}{l}{32.41} & \multicolumn{1}{l}{\underline{33.39}} & \multicolumn{1}{l}{\textbf{33.71}}\\

    & \multirow{4}{*}{DN}

    & \multicolumn{1}{l}{PA} & \multicolumn{1}{l}{63.50} & \multicolumn{1}{l}{64.06} & \multicolumn{1}{l}{63.25} & \multicolumn{1}{l}{64.51} & \multicolumn{1}{l}{\underline{64.98}} & \multicolumn{1}{l}{\textbf{65.37}}\\

    & & \multicolumn{1}{l}{mAP} & \multicolumn{1}{l}{72.61} & \multicolumn{1}{l}{73.07} & \multicolumn{1}{l}{72.15} & \multicolumn{1}{l}{73.85} & \multicolumn{1}{l}{\underline{74.02}} & \multicolumn{1}{l}{\textbf{74.67}}\\

    & & \multicolumn{1}{l}{mIoU} & \multicolumn{1}{l}{43.02} & \multicolumn{1}{l}{43.75} & \multicolumn{1}{l}{42.85} & \multicolumn{1}{l}{44.16} & \multicolumn{1}{l}{\underline{44.75}} & \multicolumn{1}{l}{\textbf{44.82}}\\
    
    & & \multicolumn{1}{l}{mF1} & \multicolumn{1}{l}{31.04} & \multicolumn{1}{l}{32.31} & \multicolumn{1}{l}{30.83} & \multicolumn{1}{l}{33.93} & \multicolumn{1}{l}{\underline{34.14}} & \multicolumn{1}{l}{\textbf{34.59}}\\
    
     \midrule
    
    \multirow{12}{*}{VOC}
    & \multirow{4}{*}{CN}

    & \multicolumn{1}{l}{PA} & \multicolumn{1}{l}{72.54} & \multicolumn{1}{l}{72.09} & \multicolumn{1}{l}{63.32} & \multicolumn{1}{l}{69.83} & \multicolumn{1}{l}{\underline{72.68}} & \multicolumn{1}{l}{\textbf{72.81}}\\

    & & \multicolumn{1}{l}{mAP} & \multicolumn{1}{l}{77.27} & \multicolumn{1}{l}{79.47} & \multicolumn{1}{l}{68.83} & \multicolumn{1}{l}{80.45} & \multicolumn{1}{l}{\underline{81.35}} & \multicolumn{1}{l}{\textbf{81.79}}\\

    & & \multicolumn{1}{l}{mIoU} & \multicolumn{1}{l}{50.28} & \multicolumn{1}{l}{50.63} & \multicolumn{1}{l}{48.86} & \multicolumn{1}{l}{49.76} & \multicolumn{1}{l}{\underline{51.12}} & \multicolumn{1}{l}{\textbf{51.29}}\\
    
    & & \multicolumn{1}{l}{mF1} & \multicolumn{1}{l}{35.24} & \multicolumn{1}{l}{35.67} & \multicolumn{1}{l}{33.26} & \multicolumn{1}{l}{34.51} & \multicolumn{1}{l}{\underline{35.92}} & \multicolumn{1}{l}{\textbf{36.57}}\\
    
    & \multirow{4}{*}{RN}

    & \multicolumn{1}{l}{PA} & \multicolumn{1}{l}{68.74} & \multicolumn{1}{l}{69.01} & \multicolumn{1}{l}{68.61} & \multicolumn{1}{l}{68.00} & \multicolumn{1}{l}{\underline{69.38}} & \multicolumn{1}{l}{\textbf{69.74}}\\

    & & \multicolumn{1}{l}{mAP} & \multicolumn{1}{l}{79.68} & \multicolumn{1}{l}{79.96} & \multicolumn{1}{l}{79.41} & \multicolumn{1}{l}{78.02} & \multicolumn{1}{l}{\underline{81.02}} & \multicolumn{1}{l}{\textbf{81.49}}\\

    & & \multicolumn{1}{l}{mIoU} & \multicolumn{1}{l}{49.44} & \multicolumn{1}{l}{49.91} & \multicolumn{1}{l}{49.15} & \multicolumn{1}{l}{49.32} & \multicolumn{1}{l}{\underline{50.43}} & \multicolumn{1}{l}{\textbf{51.58}}\\
    
    & & \multicolumn{1}{l}{mF1} & \multicolumn{1}{l}{33.08} & \multicolumn{1}{l}{33.56} & \multicolumn{1}{l}{32.69} & \multicolumn{1}{l}{32.74} & \multicolumn{1}{l}{\underline{34.28}} & \multicolumn{1}{l}{\textbf{34.68}}\\

    & \multirow{4}{*}{DN}

    & \multicolumn{1}{l}{PA} & \multicolumn{1}{l}{68.43} & \multicolumn{1}{l}{68.78} & \multicolumn{1}{l}{68.24} & \multicolumn{1}{l}{68.36} & \multicolumn{1}{l}{\underline{68.89}} & \multicolumn{1}{l}{\textbf{68.95}}\\

    & & \multicolumn{1}{l}{mAP} & \multicolumn{1}{l}{78.68} & \multicolumn{1}{l}{79.06} & \multicolumn{1}{l}{78.52} & \multicolumn{1}{l}{78.62} & \multicolumn{1}{l}{\underline{79.43}} & \multicolumn{1}{l}{\textbf{79.66}}\\

    & & \multicolumn{1}{l}{mIoU} & \multicolumn{1}{l}{49.29} & \multicolumn{1}{l}{49.68} & \multicolumn{1}{l}{49.03} & \multicolumn{1}{l}{49.11} & \multicolumn{1}{l}{\underline{49.91}} & \multicolumn{1}{l}{\textbf{50.24}}\\
    
    & & \multicolumn{1}{l}{mF1} & \multicolumn{1}{l}{32.92} & \multicolumn{1}{l}{33.83} & \multicolumn{1}{l}{32.28} & \multicolumn{1}{l}{32.56} & \multicolumn{1}{l}{\underline{34.11}} & \multicolumn{1}{l}{\textbf{34.26}}\\
    \bottomrule
  \end{tabular}}

  \end{center}
  \vspace{-2mm}

\end{table}
\begin{table}

  \vspace{2mm}
   \caption{Segmentation tests on three datasets (ViT models). For all metrics, higher is better. See Sec.~\ref{subsec:res_seg} for details.
  }
  \label{tab:vit_segmentation}
  \vspace{-3mm}
   \begin{center}

   \scalebox{1}{
    \begin{tabular}{@{}lc@{}lc@{}lc@{}lc@{}lc@{}}
    \toprule
      & & & \multicolumn{1}{l}{T-Attr} & \multicolumn{1}{l}{GAE} & \multicolumn{1}{l}{DIX2} & \multicolumn{1}{l}{DIX3}\\
    \midrule
    \multirow{8}{*}{IN-Seg}
    & \multirow{4}{*}{ViT-B}
    & \multicolumn{1}{l}{PA} & \multicolumn{1}{l}{79.70} & \multicolumn{1}{l}{76.30} & \multicolumn{1}{l}{\underline{79.91}} & \multicolumn{1}{l}{\textbf{81.02}}\\
    
    & & \multicolumn{1}{l}{mAP} & \multicolumn{1}{l}{86.03} & \multicolumn{1}{l}{85.28} &\multicolumn{1}{l}{\underline{87.12}} & \multicolumn{1}{l}{\textbf{87.45}}\\
    
    & & \multicolumn{1}{l}{mIoU} & \multicolumn{1}{l}{61.95} & \multicolumn{1}{l}{58.34} & \multicolumn{1}{l}{\underline{62.53}} & \multicolumn{1}{l}{\textbf{63.47}}\\
    
    & & \multicolumn{1}{l}{mF1} & \multicolumn{1}{l}{40.17} & \multicolumn{1}{l}{41.85} & \multicolumn{1}{l}{\underline{44.94}} & \multicolumn{1}{l}{\textbf{45.66}}\\
    
    & \multirow{4}{*}{ViT-S}
    & \multicolumn{1}{l}{PA} & \multicolumn{1}{l}{80.86} & \multicolumn{1}{l}{76.66} & \multicolumn{1}{l}{\underline{81.54}}& \multicolumn{1}{l}{\textbf{81.83}}\\
    
    & & \multicolumn{1}{l}{mAP} & \multicolumn{1}{l}{86.13} & \multicolumn{1}{l}{84.23} &
    \multicolumn{1}{l}{\underline{86.48}}& \multicolumn{1}{l}{\textbf{86.96}}\\
    
    & & \multicolumn{1}{l}{mIoU} & \multicolumn{1}{l}{63.61} & \multicolumn{1}{l}{57.70} & \multicolumn{1}{l}{\underline{64.13}} & \multicolumn{1}{l}{\textbf{64.67}}\\
    
    & & \multicolumn{1}{l}{mF1} & \multicolumn{1}{l}{43.60} & \multicolumn{1}{l}{40.72} & \multicolumn{1}{l}{\underline{46.34}} & \multicolumn{1}{l}{\textbf{46.82}}\\
    
    \midrule
    
    \multirow{8}{*}{COCO}
    & \multirow{4}{*}{ViT-B}
    & \multicolumn{1}{l}{PA} & \multicolumn{1}{l}{68.89}  & \multicolumn{1}{l}{67.10}  & \multicolumn{1}{l}{\underline{68.95}} & \multicolumn{1}{l}{\textbf{69.42}}\\
    
    & & \multicolumn{1}{l}{mAP} & \multicolumn{1}{l}{78.57} & \multicolumn{1}{l}{78.72}  &\multicolumn{1}{l}{\underline{80.63}} & \multicolumn{1}{l}{\textbf{81.22}}\\
    
    & & \multicolumn{1}{l}{mIoU} & \multicolumn{1}{l}{46.62} & \multicolumn{1}{l}{46.51} & \multicolumn{1}{l}{\underline{47.75}}  & \multicolumn{1}{l}{\textbf{47.79}}\\
    
    & & \multicolumn{1}{l}{mF1} & \multicolumn{1}{l}{26.28} & \multicolumn{1}{l}{31.70} & \multicolumn{1}{l}{\underline{33.87}} & \multicolumn{1}{l}{\textbf{34.12}}\\

    & \multirow{4}{*}{ViT-S}
    & \multicolumn{1}{l}{PA} & \multicolumn{1}{l}{69.90} & \multicolumn{1}{l}{67.95}  & \multicolumn{1}{l}{\underline{70.41}}& \multicolumn{1}{l}{\textbf{70.64}}\\
    
    & & \multicolumn{1}{l}{mAP} & \multicolumn{1}{l}{79.28} & \multicolumn{1}{l}{78.65} &\multicolumn{1}{l}{\underline{80.55}} & \multicolumn{1}{l}{\textbf{80.89}}\\
    
    & & \multicolumn{1}{l}{mIoU} & \multicolumn{1}{l}{48.62} & \multicolumn{1}{l}{46.52} & \multicolumn{1}{l}{\underline{50.81}}& \multicolumn{1}{l}{\textbf{51.22}}\\
    
    & & \multicolumn{1}{l}{mF1} & \multicolumn{1}{l}{30.88} & \multicolumn{1}{l}{30.96} & \multicolumn{1}{l}{\underline{35.61}} & \multicolumn{1}{l}{\textbf{35.74}}\\
    
     \midrule
    
    \multirow{8}{*}{VOC}
    & \multirow{4}{*}{ViT-B}
    & \multicolumn{1}{l}{PA} & \multicolumn{1}{l}{73.70} & \multicolumn{1}{l}{71.32}  & \multicolumn{1}{l}{\underline{75.33}}& \multicolumn{1}{l}{\textbf{75.84}}\\
    
    & & \multicolumn{1}{l}{mAP} & \multicolumn{1}{l}{81.08} & \multicolumn{1}{l}{80.88} &\multicolumn{1}{l}{\underline{81.75}} & \multicolumn{1}{l}{\textbf{81.89}}\\
    
    & & \multicolumn{1}{l}{mIoU} & \multicolumn{1}{l}{53.09} & \multicolumn{1}{l}{51.82} & \multicolumn{1}{l}{\underline{53.62}}  & \multicolumn{1}{l}{\textbf{53.71}}\\
    
    & & \multicolumn{1}{l}{mF1} & \multicolumn{1}{l}{31.50} & \multicolumn{1}{l}{35.72} & \multicolumn{1}{l}{\underline{36.38}} & \multicolumn{1}{l}{\textbf{36.59}}\\
    
    & \multirow{4}{*}{ViT-S}
    & \multicolumn{1}{l}{PA} & \multicolumn{1}{l}{74.96} & \multicolumn{1}{l}{71.85}  & \multicolumn{1}{l}{\underline{76.35}} & \multicolumn{1}{l}{\textbf{76.56}}\\
    
    & & \multicolumn{1}{l}{mAP} & \multicolumn{1}{l}{81.76} & \multicolumn{1}{l}{80.60} &
   \multicolumn{1}{l}{\underline{82.74}} & \multicolumn{1}{l}{\textbf{82.91}}\\
    
    & & \multicolumn{1}{l}{mIoU} & \multicolumn{1}{l}{55.37} & \multicolumn{1}{l}{51.55} & \multicolumn{1}{l}{\underline{55.83}} & \multicolumn{1}{l}{\textbf{55.98}}\\
    
    & & \multicolumn{1}{l}{mF1} & \multicolumn{1}{l}{36.03} & \multicolumn{1}{l}{34.95} & \multicolumn{1}{l}{\underline{39.27}} & \multicolumn{1}{l}{\textbf{39.41}}\\
    \bottomrule
  \end{tabular}}

  \end{center}
\end{table}

\section{Results}
\label{subsec:results}

\subsection{Explanation Tests}
\label{subsubsection:obj2}
Tables~\ref{tab:cnn_backbones_metrics_exp} and \ref{tab:appendix_vit_backbones_metrics} provide a comprehensive explanation tests for CNN and ViT models, respectively. We report results for all combinations of datasets, models, methods, and metrics.
Our analysis demonstrates that DIX consistently surpasses all baseline methods across a spectrum of metrics and architectural configurations. On CNN-based DIX variations (Tab.~\ref{tab:cnn_backbones_metrics_exp}), DIX3 outclasses DIX2 in terms of NEG, INS, SIC, and AIC metrics for both RN and DN backbones, while demonstrating dominance across all metrics for the CN backbone. 
Regarding the ViT-based DIX variants (Tab.~\ref{tab:appendix_vit_backbones_metrics}), DIX3 outperforms DIX2 across all metrics (with the exception of PIC on ViT-B, and PIC and ADP on ViT-S).These trends showcase the advantage of aggregating information from more layers.
In the context of CNNs, the second-best performing methods are GC and GC++, which leverage both activation and gradients to outperform other approaches across most evaluation metrics.
Additionally, we note that path integration techniques (IG, BIG, and GIG) demonstrate competitive results in terms of POS and DEL metrics, while displaying comparatively weaker performance in other aspects. This disparity may be attributed to the grainy output maps generated by path integration techniques, as evidenced in Fig.\ref{fig:DIXablation} for IG explanation maps on CNNs. These methods ignore the activations and integrate on the image domain only, hence missing some of the key features. This is particularly evident in the significant contrast between their strong performance on POS and the corresponding weaker performance on NEG. As path integration methods produce sparse maps that can negatively affect performance in certain metrics, , we extend our analysis to encompass the SIC and AIC metrics as well~\cite{kapishnikov2019xrai}. These metrics were originally employed to assess GIG\cite{kapishnikov2021guided} and BIG\cite{xu2020attribution}. Yet, the incorporation of SIC and AIC did not alter the trend of the results. This suggests that DIX is highly effective for generating high-quality explanation maps.
Finally, we present an ablation study in Section~\ref{sec:ablation-study}, aimed at comparing diverse versions and alternatives of DIX. 
This analysis serves to emphasize the effectiveness of the integration process and the strategic utilization of information from multiple layers within the DIX methodology.

\subsection{Segmentation Tests}
\label{subsec:res_seg}
Tables~\ref{tab:cnns_segmentation_table} and \ref{tab:vit_segmentation} present segmentation tests results on CNN and ViT models, respectively. The results are reported for all combinations of datasets, models, explanation methods, and segmentation metrics. In these experiments, only the 5 best performing CNN explanation methods from Tab.~\ref{tab:cnn_backbones_metrics_exp} are considered.
Once again, it becomes evident that DIX consistently delivers the most favorable segmentation outcomes for both CNN and ViT models. This outcome can be rationalized by the localized and precise maps that DIX generates.

% \vspace{-3mm}
\subsection{Qualitative Evaluation}
\label{subsubsection:qualitiveee}
\begin{figure}
    \centering
    \includegraphics[width=0.52\textwidth]{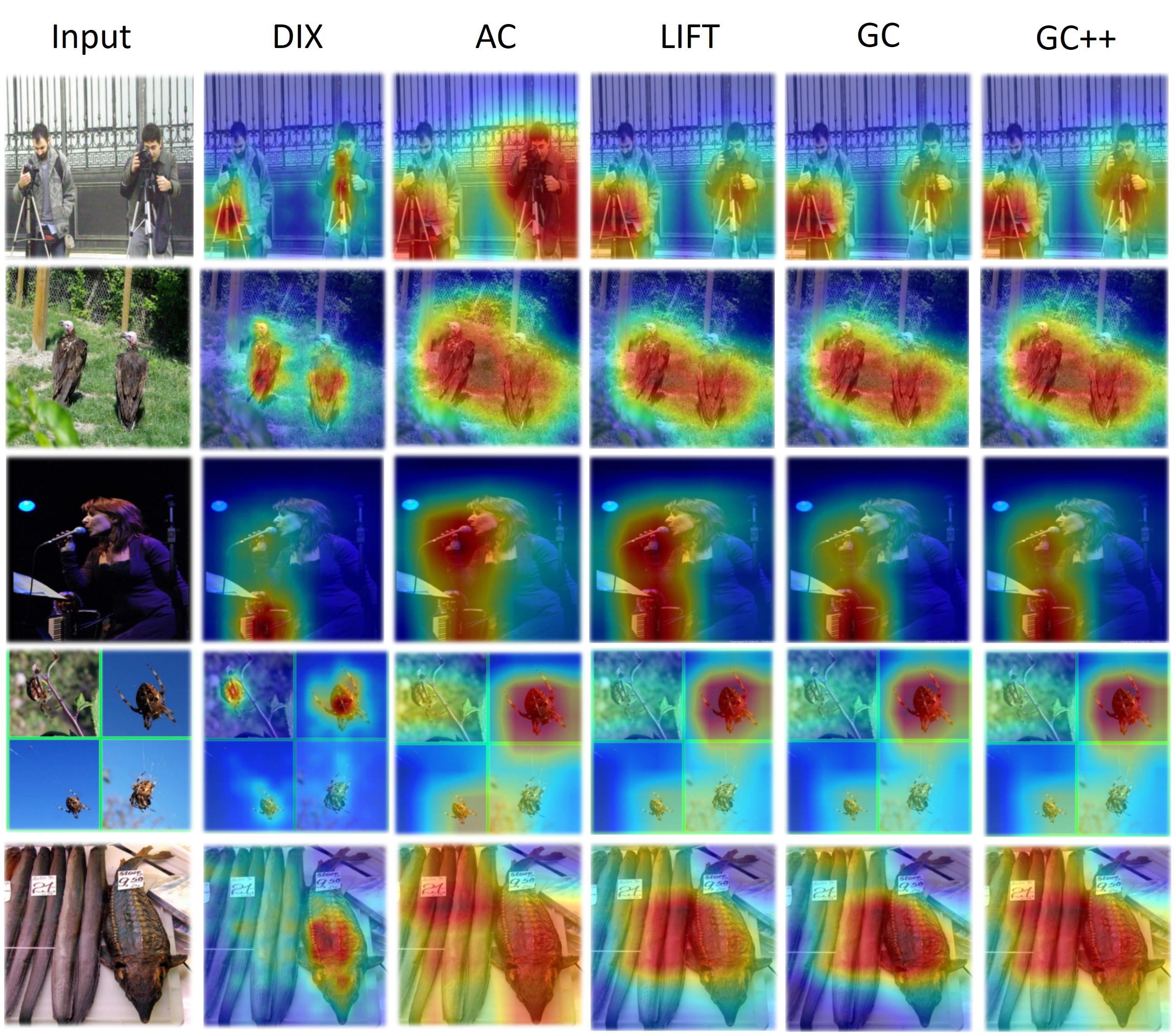}\vspace{-1mm}
    \caption{CNN Qualitative Results: Explanation maps produced using RN w.r.t. the classes (top to bottom): `tripod', `vulture', `accordion, squeeze box', `garden spider, Aranea diademata', and `sturgeon'.
    }
    \label{fig:qualitative_fig}
\end{figure}

\begin{figure}
\centering
    \includegraphics[width=0.51\textwidth]{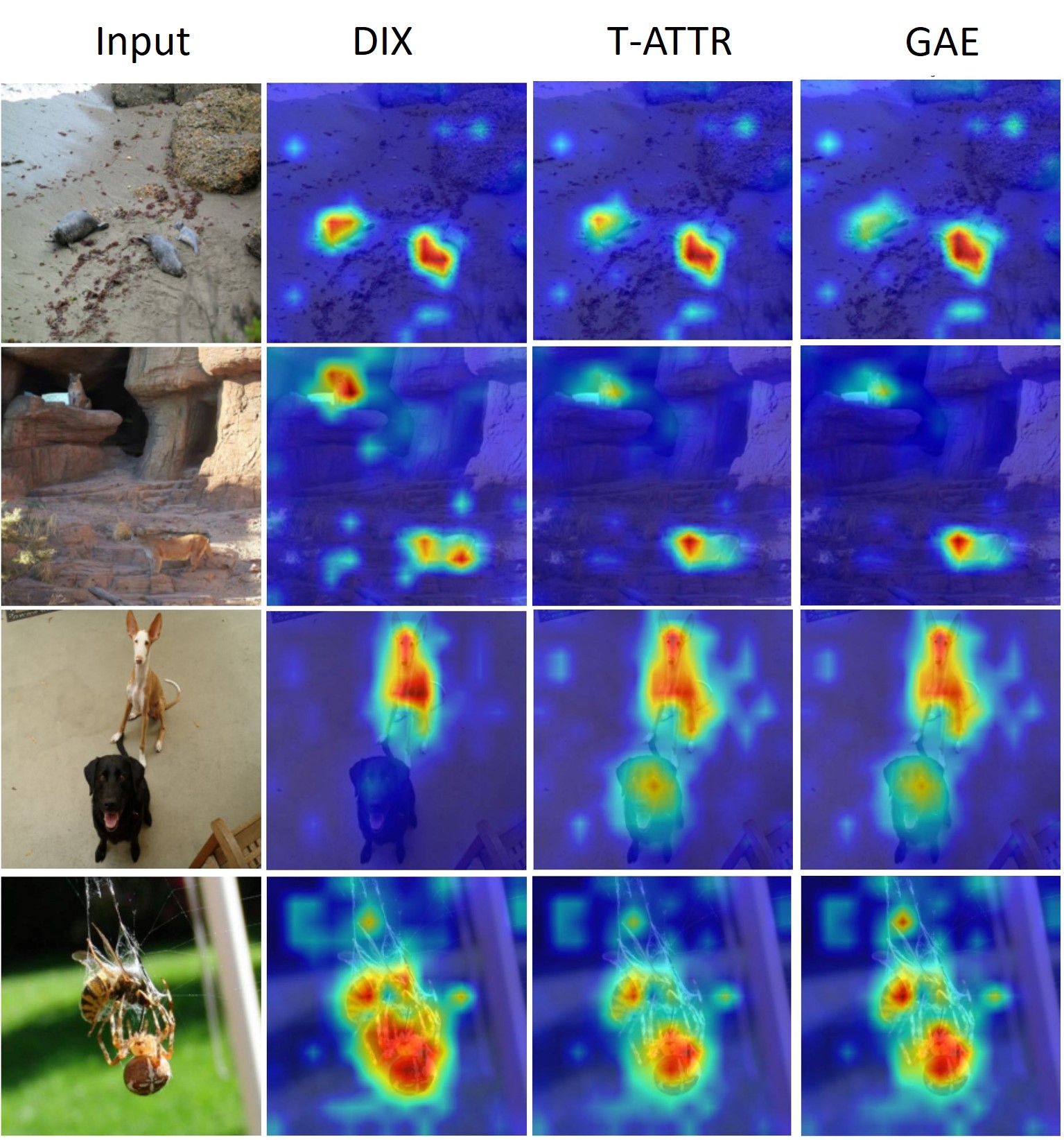}
    \caption{ViT Qualitative Results: Explanation maps produced using ViT-B w.r.t. the classes (top to bottom): `sea lion', `cougar, puma, catamount, mountain lion, painter, panther, Felis concolor', `Ibizan hound, Ibizan Podenco', and `garden spider, Aranea diademata'.}\vspace{-4mm}
    \label{fig:vitqr}
\end{figure}

Figure~\ref{fig:qualitative_fig} presents a qualitative comparison of the explanation maps obtained by the top-performing CNN explanation methods on a large set of examples that are randomly drawn from multiple classes from the IN dataset. Arguably, DIX (DIX3) produces the most accurate explanation maps in terms of class discrimination and localization. These results correlate well with the trends from Tabs.~\ref{tab:cnn_backbones_metrics_exp} and \ref{tab:cnns_segmentation_table}. We observe that in the case of class `accordion, piano accordion, and squeeze box', DIX focuses mostly on the correct item, while the gradient-free methods focus mostly on other parts of the image, exposing their class-agnostic behavior. Moreover, a similar trend is observed with the 'sturgeon' class, in which DIX is the only one to focus on the relevant class. Figure~\ref{fig:vitqr} presents a qualitative comparison of the explanation maps obtained by explanation methods for ViT. Once again, we see that DIX produces the most accurate and focused explanation maps.

\subsection{Ablation Study}
\label{sec:ablation-study}
In this work we present and evaluate DIX2 and DIX3. In this section, we justify these choices via an ablation study.
To this end, we set $n=10$, and consider three alternatives: (1) \textbf{DIX1} - we use the last layer as the only layer to interpolate i.e., $S=\{L\}$.
(2) \textbf{DIX2-MUL} - $\m^L_{\text{DIX}}$ and $\m^{L-1}_{\text{DIX}}$ are being element-wise multiplied to produce the final explanation map.
(3) \textbf{DIX3-GRADS} - we use the plain gradients without explicitly incorporating the information from the activation or attention maps.

\begin{figure*}
\centering
    \includegraphics[width=0.8\textwidth]{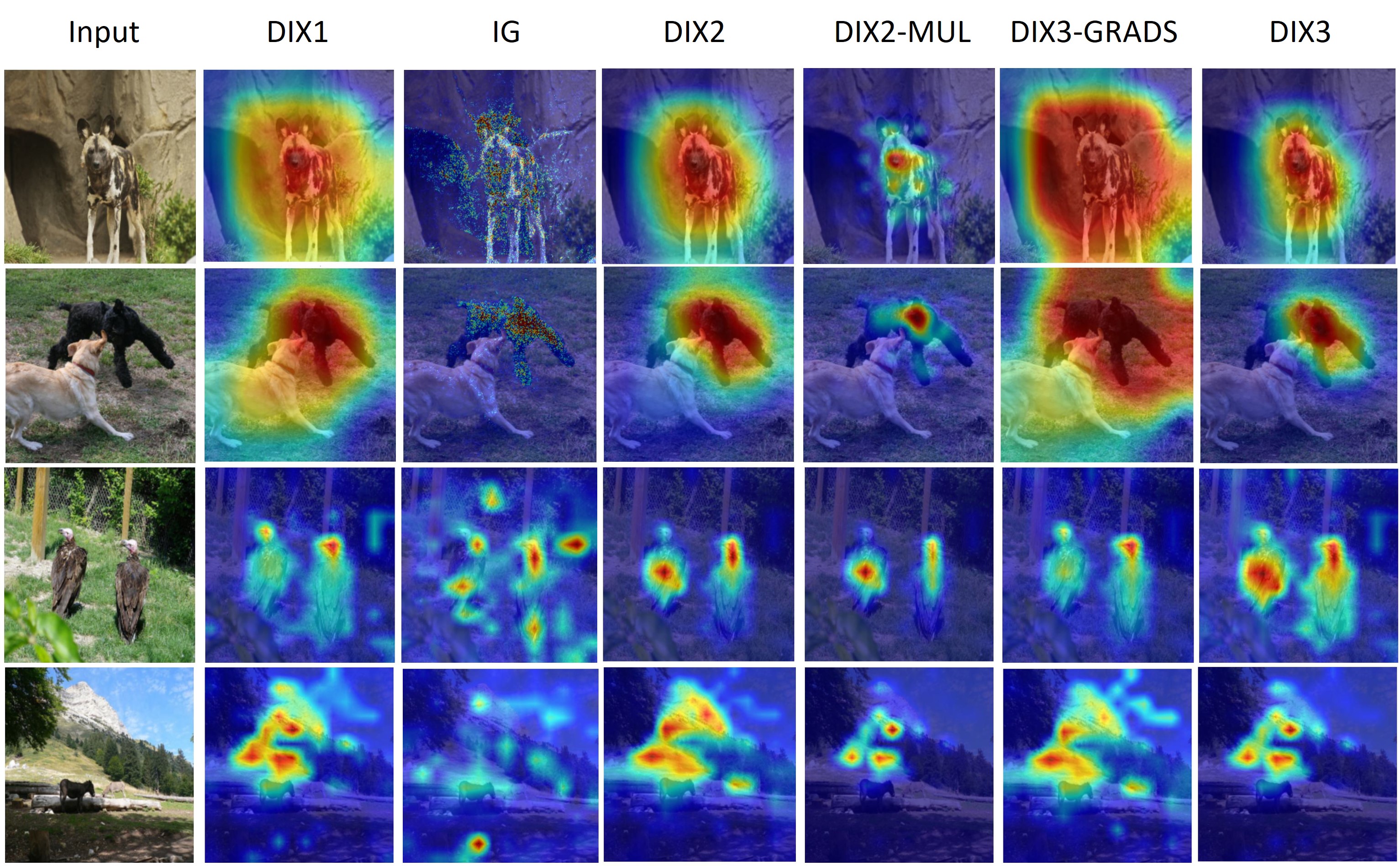}\vspace{-1mm}
    \caption{Ablation study results. Explanation maps produced using RN (rows 1,2) and ViT-B (rows 3,4) w.r.t. the classes (top to bottom): `African hunting dog, hyena dog, Cape hunting dog, Lycaon pictus', 'Kerry blue terrier', 'vulture', 'alp'.
    }\vspace{-1mm}
    \label{fig:DIXablation}
\end{figure*}

Table~\ref{tab:appendix_vit_ablation} reports the results for the RN and ViT-B models on the IN dataset. For the sake of completeness, we further include the results for IG, DIX2, and DIX3 (taken from Tabs.~\ref{tab:cnn_backbones_metrics_exp} and \ref{tab:appendix_vit_backbones_metrics}). 
First, we can see the superior performance of DIX2 and DIX3 across all metrics and models.  We further observe that both DIX1 and DIX2-MUL fall short in comparison to DIX2. This observation underscores the inherent necessity of incorporating information from additional layers and shows the advantages of aggregation via summation. 
When aggregating the explanation maps of different layers, the objective is to effectively incorporate data from each map to capture a richer spectrum of insights and class-specific signals. Notably, the multiplication operator exhibits a behavior akin to intersection, where both high pixel values are required for proper appearance in the final map. This characteristic, as depicted in Figure~\ref{fig:DIXablation}, contrasts with the intended outcome.
Furthermore, the superiority of DIX3 over DIX3-GRADS underscores the benefit from exploiting intermediate representation information alongside its corresponding gradients, which contributes to the generation of localized, accurate and class discriminative explanation maps.
The results presented in Table~\ref{tab:appendix_vit_ablation} highlight a distinct advantage for IG and DIX2-MUL with respect to the POS and DEL metrics when compared to DIX3-GRADS and DIX1, both of which generate less concentrated explanation maps. This is due to the fact that the deletion of the most relevant pixels results in fewer pixels being removed, and the
mask is more focused on a subset of pixels. DIX1, for instance, produces less focused explanation maps that may highlight irrelevant areas. Such
coarse highlighting leads to a slower decrease in the prediction score
during the deletion process.
On the contrary, DIX1 and DIX3-GRADS exhibit superior performance in relation to the NEG and INS metrics. This divergence in performance can be attributed to the expansive nature of their explanation map, resulting in numerous pixels that require removal. In the context of the NEG metric, this characteristic contributes to a slow decrease in the prediction score during the deletion process and, subsequently, a larger area under the curve (AUC).

\begin{table}[t!]
  \caption{Ablation study results for various DIX configurations on the IN dataset. See Sec.~\ref{sec:ablation-study} for details.
  }\vspace{-2mm}
  \label{tab:appendix_vit_ablation}
  \begin{center}
  \begin{small}
  \scalebox{0.98}{
    \begin{tabular}{@{}lc@{}lc@{}lc@{}lc@{}}
    \toprule
      & & \multicolumn{1}{l}{DIX1} & \multicolumn{1}{l}{IG} & \multicolumn{1}{l}{DIX2} & \multicolumn{1}{l}{DIX2-MUL} & \multicolumn{1}{l}{DIX3-GRADS} & \multicolumn{1}{l}{DIX3} \\
    \midrule
    \multirow{8}{*}{RN}

       & \multirow{1}{*}{NEG}  & \multicolumn{1}{l}{55.47} & \multicolumn{1}{l}{45.66} & \multicolumn{1}{l}{\underline{56.28}} & \multicolumn{1}{l}{55.24}  & \multicolumn{1}{l}{56.05} & \multicolumn{1}{l}{\textbf{57.13}}\\

        & \multirow{1}{*}{POS} & \multicolumn{1}{l}{17.47} & \multicolumn{1}{l}{17.24} & \multicolumn{1}{l}{\textbf{15.69}} & \multicolumn{1}{l}{17.28}  & \multicolumn{1}{l}{18.13} & \multicolumn{1}{l}{\underline{17.11}}\\

        & \multirow{1}{*}{INS}  & \multicolumn{1}{l}{47.53} & \multicolumn{1}{l}{39.87} & \multicolumn{1}{l}{\underline{48.09}} & \multicolumn{1}{l}{47.13}  &  \multicolumn{1}{l}{47.88} & \multicolumn{1}{l}{\textbf{48.91}}\\

        & \multirow{1}{*}{DEL} & \multicolumn{1}{l}{13.72} & \multicolumn{1}{l}{13.49} & \multicolumn{1}{l}{\textbf{12.84}} & \multicolumn{1}{l}{13.59}  &  \multicolumn{1}{l}{14.52}& \multicolumn{1}{l}{\underline{13.36}}\\

         & \multirow{1}{*}{ADP} & \multicolumn{1}{l}{17.21} & \multicolumn{1}{l}{37.52} & \multicolumn{1}{l}{\textbf{15.68}} & \multicolumn{1}{l}{21.38}  &  \multicolumn{1}{l}{17.43} & \multicolumn{1}{l}{\underline{16.02}}\\

        & \multirow{1}{*}{PIC}   & \multicolumn{1}{l}{36.54} & \multicolumn{1}{l}{19.94} & \multicolumn{1}{l}{\textbf{40.21}} & \multicolumn{1}{l}{28.46}  &   \multicolumn{1}{l}{37.10} & \multicolumn{1}{l}{\underline{37.29}} \\

        & \multirow{1}{*}{SIC}   & \multicolumn{1}{l}{76.85} & \multicolumn{1}{l}{54.67} & \multicolumn{1}{l}{\underline{77.61}} & \multicolumn{1}{l}{75.17}  &  \multicolumn{1}{l}{76.13} & \multicolumn{1}{l}{\textbf{78.12}}\\

        & \multirow{1}{*}{AIC}  & \multicolumn{1}{l}{75.48} & \multicolumn{1}{l}{51.92} & \multicolumn{1}{l}{\underline{76.09}} & \multicolumn{1}{l}{74.21}  &  \multicolumn{1}{l}{74.88} & \multicolumn{1}{l}{\textbf{76.34}}\\
           
        \midrule

    \multirow{8}{*}{ViT-B}
    
        & \multirow{1}{*}{NEG}  & \multicolumn{1}{l}{55.98} & \multicolumn{1}{l}{40.94} & \multicolumn{1}{l}{\underline{56.43}} & \multicolumn{1}{l}{55.62}  &  \multicolumn{1}{l}{55.78} & \multicolumn{1}{l}{\textbf{56.94}}\\

        & \multirow{1}{*}{POS}  & \multicolumn{1}{l}{15.49} & \multicolumn{1}{l}{22.43} & \multicolumn{1}{l}{\underline{15.10}} & \multicolumn{1}{l}{15.37}  &  \multicolumn{1}{l}{{15.81}} & \multicolumn{1}{l}{\textbf{14.85}}\\

        & \multirow{1}{*}{INS} & \multicolumn{1}{l}{49.38} & \multicolumn{1}{l}{35.07} & \multicolumn{1}{l}{\underline{49.51}} & \multicolumn{1}{l}{49.27}  &  \multicolumn{1}{l}{49.33} & \multicolumn{1}{l}{\textbf{50.59}}\\

        & \multirow{1}{*}{DEL}  & \multicolumn{1}{l}{13.06} & \multicolumn{1}{l}{17.90} & \multicolumn{1}{l}{\underline{12.62}} & \multicolumn{1}{l}{12.85}  &  \multicolumn{1}{l}{{13.12}}& \multicolumn{1}{l}{\textbf{12.16}}\\

         & \multirow{1}{*}{ADP} & \multicolumn{1}{l}{36.96} & \multicolumn{1}{l}{41.35} & \multicolumn{1}{l}{\underline{35.93}} & \multicolumn{1}{l}{38.62}  &  \multicolumn{1}{l}{{36.08}} & \multicolumn{1}{l}{\underline{35.58}}\\

        & \multirow{1}{*}{PIC}  & \multicolumn{1}{l}{26.94} & \multicolumn{1}{l}{16.89} & \multicolumn{1}{l}{\textbf{28.21}} & \multicolumn{1}{l}{26.39}  &  \multicolumn{1}{l}{{27.13}} & \multicolumn{1}{l}{\textbf{27.41}}\\

        & \multirow{1}{*}{SIC}   & \multicolumn{1}{l}{67.79} & \multicolumn{1}{l}{58.91} & \multicolumn{1}{l}{\underline{68.94}} & \multicolumn{1}{l}{68.43}  &  \multicolumn{1}{l}{{68.32}} & \multicolumn{1}{l}{\textbf{69.11}}\\

        & \multirow{1}{*}{AIC}  & \multicolumn{1}{l}{61.56} & \multicolumn{1}{l}{54.93} & \multicolumn{1}{l}{\underline{62.42}} & \multicolumn{1}{l}{62.18}  &  \multicolumn{1}{l}{{61.94}} & \multicolumn{1}{l}{\textbf{65.03}}\\
        \bottomrule
  \end{tabular}}
  \end{small}
  \end{center}
\end{table}

\subsection{Sanity Tests}
\label{subsec:res_sanity}
In what follows, we show that DIX passes all sanity tests successfully, thereby furnishing additional substantiation for the authenticity of DIX as a robust machinery for generating accurate explanation maps.

\begin{figure}
    \centering
    \includegraphics[scale=0.5]{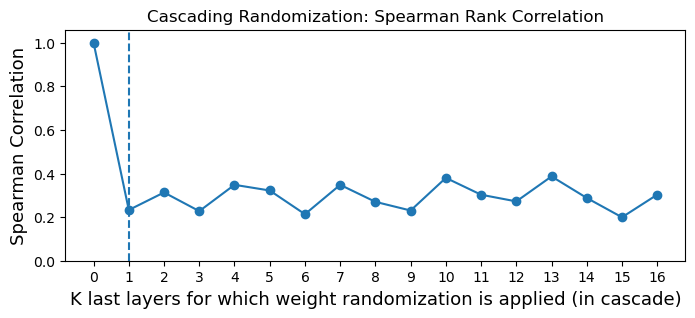}
    \vspace{-7mm}
    \caption{\textbf{Cascading Randomization}: 
    The presented graph depicts the Spearman rank correlation (averaged on 50K examples) between the explanation produced by DIX using the original and randomized model's weights. The x-axis corresponds to the number of layers being randomized, starting from the output layer. The dashed line indicates the point where the successive randomization of the network commences, which is at the top layer. The first dot (x=0) corresponds to no randomization (the original model is used), yielding perfect correlation between the explanation maps.
    }%\vspace{-3mm}
    \label{fig:sanity_checkes_metric_cascade}
    
\end{figure}

\paragraph{Cascading Randomization}
Figure~\ref{fig:sanity_checkes_metric_cascade} shows the Spearman correlation, computed as an average across 50K examples, between the initial explanation map derived from DIX and the original (pretrained) VGG-19 model, as well as the explanation map obtained from DIX and each of the cascade randomization variations of the original model. The markers on the x-axis are between '0' and '16', where $x=k$ means that the weights of the last $k$ layers of the model are randomized.
Notably, at $x=0$, there is no randomization, hence the correlation with the original model is perfect. Starting from $x=1$ (marked by the horizontal dashed line) and up to $x=16$, the graph depicts a progressive cascade randomization of the original model. It is evident that, with an increase in the randomization of layer weights, the correlation with the explanation map of the original model experiences a significant decline. This trend underscores the sensitivity of DIX to the parameters of the model, a characteristic that is both anticipated and desirable for any explanation approach, as emphasized by~\cite{adebayo2018sanity}.

\begin{figure}[t]
    \centering
    \includegraphics[scale=0.5]{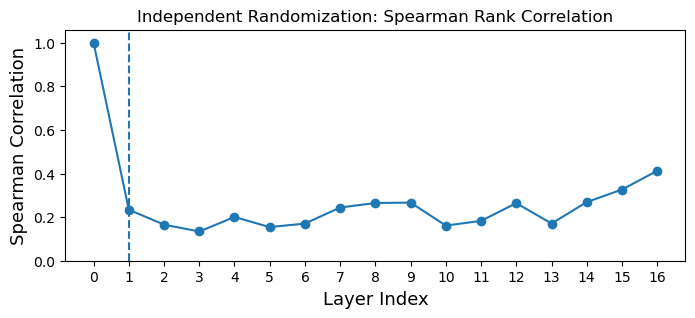}
    \vspace{-7mm}
    \caption{\textbf{Independent Randomization}: 
    The y-axis of the presented graph represents the rank correlation between the original and randomized explanations, with each point on the x-axis corresponding to a specific layer of the model. The dashed line marks the point where the randomization of the network layers commences, which is at the top layer.
    }
\label{fig:sanity_checkes_metric_ind}
\end{figure}

% \vspace{-1mm}
\paragraph{Independent Randomization}
Figure~\ref{fig:sanity_checkes_metric_ind} presents results for the independent randomization tests.
For $x=0$, no randomization was introduced and the correlation to the original model is perfect. As $x$ advances to $i$ ($i>0$), the graph illustrates the correlation of the initial model with a configuration where solely the weights of the $i$-th penultimate layer were randomized, while the weights of all other layers remained the same. Evidently, the correlation values exhibit a consistent diminution across all layers, underscoring DIX's sensitivity to weight randomization on an individual layer basis. This characteristic is fundamentally desirable for an explanation technique, serving as evidence of its sensitivity to the distinct layers within the model.

\begin{figure}
    \centering
    \includegraphics[scale=0.38]{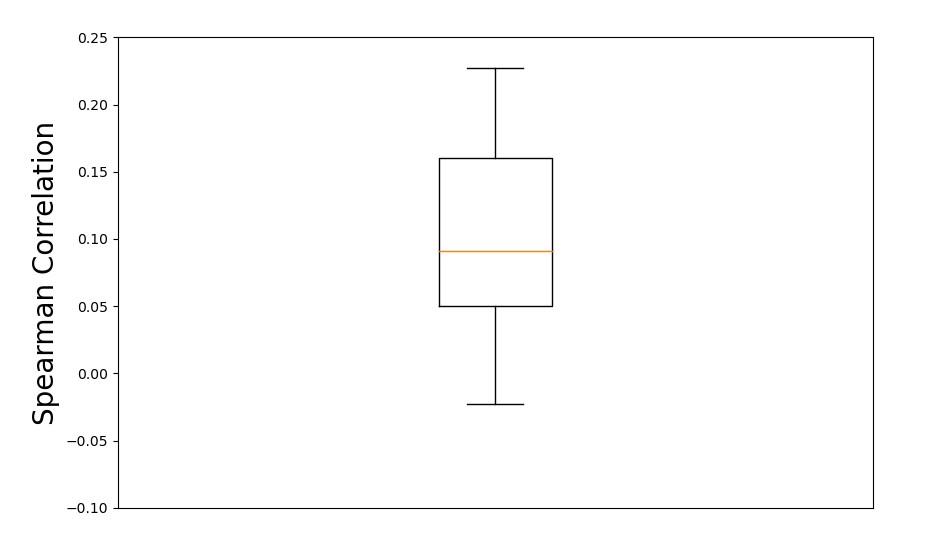}
    \vspace{-3mm}
    \caption{\textbf{Data Randomization Test}: Spearman rank correlation box plot for DIX with the VGG-19 model.}
    \label{fig:sanity_checkes_visual_bxplt}
    \vspace{-5mm}
\end{figure}

\paragraph{Data Randomization}
Figure~\ref{fig:sanity_checkes_visual_bxplt} presents a box plot computed for the Spearman correlation values obtained for paired explanation maps (50K examples): one produced using the original model that is trained with the ground truth, and another produced by the model trained with the permuted labels. 
We can see that the correlation values are very low indicating DIX's sensitivity to the labeling of the dataset. Hence, we conclude that DIX successfully passes the data randomization test.

\section{Conclusion}
\label{sec:conclusion}
We presented the Deep Integrated Explanations (DIX) method for producing explanations for vision models.
DIX is founded upon the accumulation of maps originating from multiple layers, encompassing interpolated network representations along with their corresponding gradients. We demonstrated the applicability of DIX for explaining CNN and ViT models, where it is shown to outperform state-of-the-art explanation methods across multiple tasks, datasets, network architectures, and metrics. Finally, we validated DIX as a machinery for generating faithful explanation maps via an extensive set of sanity tests.
\clearpage

\bibliographystyle{cikm23.bst}
\balance
\bibliography{cikm23}

\end{document}